\documentclass[11pt, a4paper, logo, twocolumn]{rai}
    \usepackage[toc,page]{appendix}
\usepackage{svg}
\usepackage{CJKutf8}
\usepackage{tcolorbox}
\usepackage{multirow}
\usepackage{multicol}
\usepackage{tablefootnote} %
\usepackage{subcaption}
\usepackage{graphicx}
\usepackage{float}
\usepackage{amssymb}
\usepackage{amsmath}

\DeclareMathOperator*{\argmin}{argmin}
\usepackage{array}
\newcolumntype{C}{>{\centering\arraybackslash}X}
\usepackage[capitalize]{cleveref}
\crefname{section}{Sec.}{Secs.}
\Crefname{section}{Section}{Sections}
\Crefname{table}{Table}{Tables}
\crefname{table}{Tab.}{Tabs.}

\usepackage{mathtools}

\def\eg{\emph{e.g.}}

\usepackage{tikz}

\usepackage{pifont}%

\usepackage{soul}
\usepackage{caption} %

\usepackage[textsize=tiny]{todonotes}
\newcounter{todocounter}

\usepackage{xspace}
\usepackage{tabularx}
\usepackage{adjustbox}
\newcommand{\autosizeTable}[1]{%
    \begin{adjustbox}{max width=\linewidth}
        #1
    \end{adjustbox}
}

\definecolor{Gray}{gray}{0.9}
\definecolor{LightGray}{gray}{0.7}
\newcolumntype{g}{>{\color{LightGray}}r}
\newcommand{\shortname}{NovaFlow\xspace}

\definecolor{LightGray}{gray}{0.7}
\definecolor{gold}{RGB}{255, 199, 44}
\definecolor{purple}{RGB}{174, 78, 255}
\colorlet{gold}{gold!30!white}
\colorlet{purple}{purple!50!white}
\newcolumntype{g}{>{\color{LightGray}}r}

\begin{document}
\runningtitle{\shortname: Zero-Shot Manipulation via Actionable Flow from Generated Videos}

\title{\LARGE \bf
\shortname: Zero-Shot Manipulation via \\ Actionable Flow from Generated Videos
}

\author[1,2*]{Hongyu Li}
\author[1*]{Lingfeng Sun}
\author[1]{Yafei Hu}
\author[1]{Duy Ta}
\author[1]{Jennifer Barry}
\author[2]{George Konidaris}
\author[1]{Jiahui Fu}

    \affil[*]{Equal contribution} 
    \affil[1]{Robotics and AI Institute} 
    \affil[2]{Brown University}
    \correspondingauthor{
    Hongyu Li, George Konidaris (\href{mailto:hli230@cs.brown.edu,gdk@cs.brown.edu}{\{hli230,gdk\}@cs.brown.edu}),
    Lingfeng Sun, and
    Jiahui Fu (\href{mailto:lsun@rai-inst.com,jfu@rai-inst.com}{\{lsun, jfu\}@rai-inst.com}). \\
    Hongyu Li and George Konidaris were supported by the Office of Naval Research (ONR) under REPRISM MURI N000142412603 and ONR grant N00014-22-1-2592. Partial funding was also provided by the Robotics and AI Institute.
    }

\begin{abstract}
Enabling robots to execute novel manipulation tasks zero-shot is a central goal in robotics. Most existing methods assume in-distribution tasks or rely on fine-tuning with embodiment-matched data, limiting transfer across platforms. We present \shortname, an autonomous manipulation framework that converts a task description into an actionable plan for a target robot without any demonstrations. Given a task description, \shortname synthesizes a video using a video generation model and distills it into 3D actionable object flow using off-the-shelf perception modules. From the object flow, it computes relative poses for rigid objects and realizes them as robot actions via grasp proposals and trajectory optimization. For deformable objects, this flow serves as a tracking objective for model-based planning with a particle-based dynamics model. By decoupling task understanding from low-level control, \shortname naturally transfers across embodiments. We validate on rigid, articulated, and deformable object manipulation tasks using a table-top Franka arm and a Spot quadrupedal mobile robot, and achieve effective zero-shot execution without demonstrations or embodiment-specific training. Project website: \url{https://novaflow.lhy.xyz/}.
\end{abstract}

\twocolumn[{%
\renewcommand\twocolumn[1][]{#1}%
\maketitle
\begin{center}
    \centering
    \captionsetup{type=figure}
     \includegraphics[width=\textwidth]{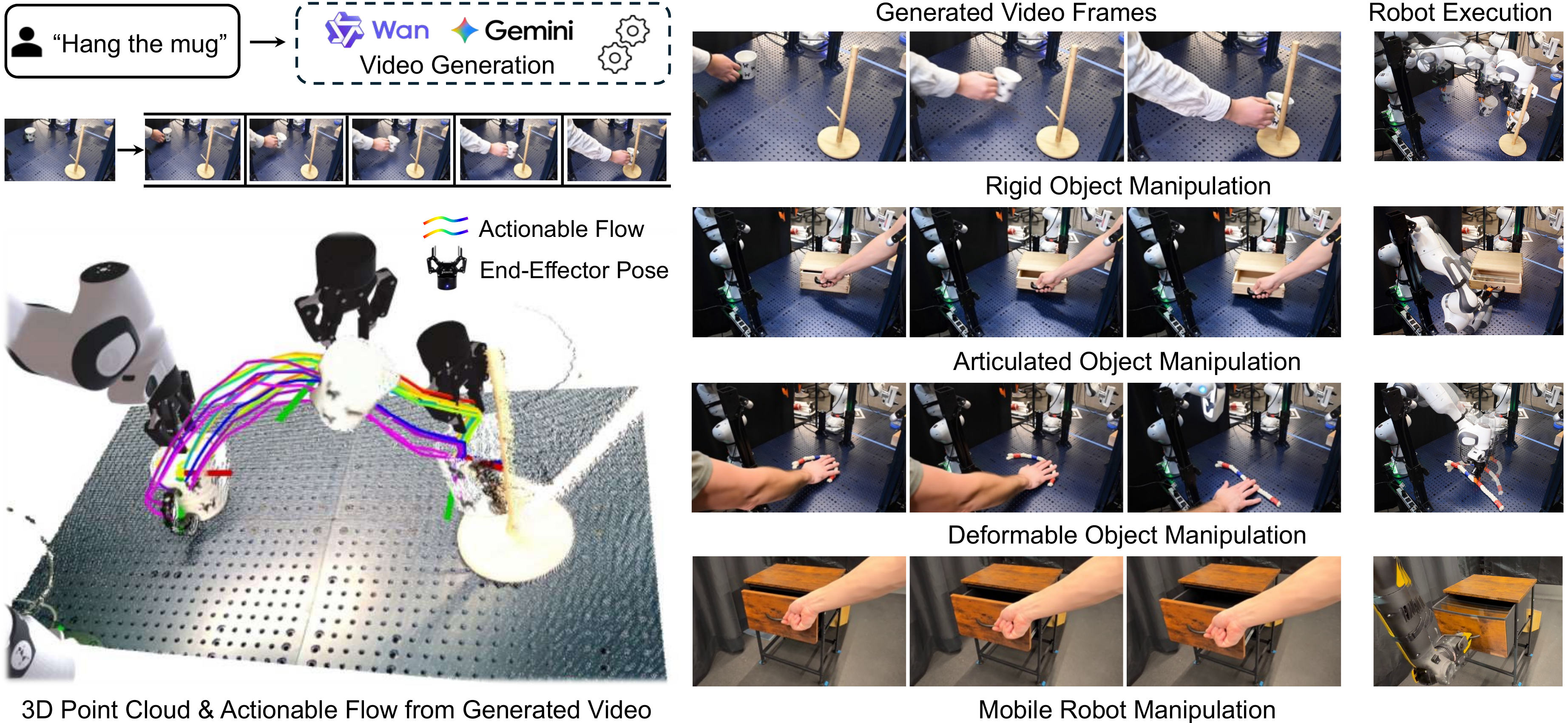}
    \caption{
        \textbf{\shortname manipulation framework. } A generated task-solving video  is distilled into a \textbf{3D actionable object flow} aligned with the robot's observation. From this flow, reference end-effector trajectories are computed and tracked, enabling robots to manipulate rigid, articulated, and deformable objects across different embodiments \emph{without demonstrations}.
    }
    \label{fig:teaser}
\end{center}
}]

\section{Introduction}
A long-standing goal in robotics is to build generalist robots capable of performing a wide variety of manipulation tasks in unstructured environments without task-specific training.
Many believe that Vision-Language-Action (VLA) models~\cite{kim_openvla_2025,brohan_rt-1_2022,ghosh_octo_2024,team_careful_2025} can achieve this generalization, following the success of Large Language Models (LLMs)~\cite{qwen_qwen25_2025,deepseek-ai_deepseek-r1_2025,touvron_llama_2023}, Vision-Language Models (VLMs)~\cite{liu_visual_2023,openai_gpt-4o_2024}, and video generation models~\cite{blattmann_stable_2023,yang_cogvideox_2024,kong_hunyuanvideo_2025,wan_wan_2025} that learn from vast, internet-scale datasets. 
However, directly applying this paradigm to robotics creates a significant data bottleneck. 
VLA models, for their end-to-end training, require vast quantities of robot-specific vision-language-action data that is difficult and expensive to collect, a stark contrast to the readily available web-scale data used for LLMs and VLMs.

An alternative path towards generalist robots lies in creating modular systems that decompose the problem into task understanding and robot control.
These systems leverage powerful pretrained models~\cite{stone_open-world_2023,dalal_local_2025} and traditional robotic engineering methods like inverse kinematics (IK)~\cite{patel_robotic_2025} or model predictive control~\cite{yu_language_2023} to bypass large-scale robot data collection, a promising strategy for closing the data gap~\cite{goldberg_good_2025}. 
For instance, some approaches use large language or vision-language models to generate high-level plans, affordance maps, or semantic keypoints to guide the robot~\cite{black_zero-shot_2023,bharadhwaj_towards_2024,huang_rekep_2025,yin_codediffuser_2025}. 
While these methods successfully offload semantic reasoning to large models, translating this understanding into physical actions remains an open problem.
The control policy, for instance, relies on either predefined skill primitives (\eg, opening a drawer)~\cite{liang_code_2023,dalal_local_2025} or learned skills from real-world demonstrations~\cite{black_zero-shot_2023,xu_flow_2025,bharadhwaj_towards_2024, stone_open-world_2023,yin_codediffuser_2025,huang_imagination_2024}.
This approach reintroduces the data bottleneck and limits generalizability and scalability.

To overcome these limitations, we propose \textbf{\shortname}, a novel framework that breaks the dependency on robot data to achieve autonomous manipulation. 
Our key insight is to \emph{repurpose large-scale pretrained video generation models as a source of commonsense task understanding and implicit physical knowledge for deriving object motion}. 
We hypothesize that by training on internet-scale video data, these models have already captured a rich, generalizable understanding of task and object dynamics that can be leveraged for unseen objects, environments, and tasks. 
This separates our approach from prior work that relies on self-collected data to train smaller, specialized video models~\cite{du_learning_2023,liang_dreamitate_2025,li_unified_2025,ko_learning_2023}.
To translate this understanding from video to robot actions, we leverage  \emph{actionable 3D object flow}, a generalized atomic representation of object motion.

\shortname generates robot actions from a single visual observation and task description and consists of two components: a flow generator and a flow executor. 
The flow generator leverages large-scale video generation models to distill generalized knowledge of object motion into an \emph{actionable 3D object flow}.
This is achieved using a pipeline of pretrained perception modules for monocular depth estimation~\cite{li_megasam_2025}, 3D point tracking~\cite{zhang_tapip3d_2025}, and object grounding~\cite{liu_grounding_2025, ravi_sam_2024}.
The flow executor then translates this 3D flow into robot actions using IK and trajectory optimization, requiring no robot-specific data or task training. 
To handle diverse object types, the executor uses correspondence-based model-free tracking for rigid and articulated objects~\cite{kabsch_solution_1976} and dynamic model-based planning with particle models for deformable objects~\cite{jiang_phystwin_2025,zhang_particle-grid_2025}, using the flow as a tracking objective. 

In summary, we present \shortname, an object-centric and embodiment-agnostic framework for autonomous manipulation that requires no task-specific tuning. 
We demonstrate its efficacy across both tabletop and mobile manipulator tasks involving rigid, articulated, and deformable objects. 
At its core, we introduce an actionable 3D object flow representation that is key to its generalizability and achieve state-of-the-art zero-shot performance on a range of real-world tasks, outperforming previous demonstration-free and data-dependent methods. 

\begin{figure*}
    \centering
    \includegraphics[width=\textwidth]{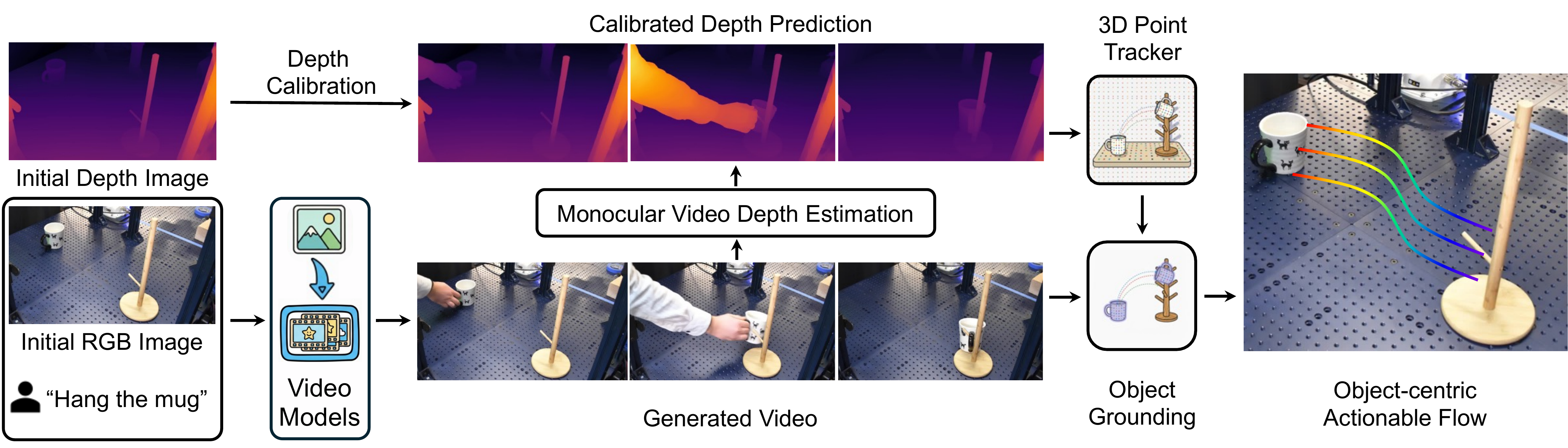}
    \caption{
        \textbf{Flow generator pipeline.}
        Given an initial image and a task prompt,  a video model is used to generate a video of the plausible object motion. This video is then processed by pretrained perception modules to distill an actionable 3D object flow. This involves (1) lifting the 2D video to 3D using monocular depth estimation, 
        (2) calibrating the estimated depth against the initial depth, 
        (3) tracking the dense per-point motion using 3D point tracking, 
        and (4) extracting the object-centric 3D flow via object grounding.
    }
    \vspace{-1em}
    \label{fig:flow-generator}
\end{figure*}

\section{Related Work}
We define an approach as zero-shot or demonstration-free if it does not require collecting any robot-specific data or task-specific training.
While LLMs and VLMs have shown promising zero-shot capabilities, their embodied successors, VLA models, have yet to achieve the same level of generalization.
Recent VLAs~\cite{lee_molmoact_2025,black_pi_0_2024} still rely on data collection to generalize on novel embodiments or camera views.
This is due to the data bottleneck created by the end-to-end training nature of VLAs.
To address this, we decouple the task understanding (Sec.~\ref{sec:video-based-manipulation}) and robot control, bridged by an intermediate representation, the 3D object flow (Sec.~\ref{sec:flow-based-manipulation}).

\subsection{Video-based Manipulation}
\label{sec:video-based-manipulation}
Prior work has utilized video generation models for manipulation.
Video can be a generalized representation of motion, which serves as a visual instruction for robots to execute tasks.
Some work trains an inverse dynamics model~\cite{du_learning_2023,ajay_compositional_2023} or a policy~\cite{bharadhwaj_gen2act_2024,li_unified_2025,liang_video_2025} to convert the generated video into robot actions.
Other work tracks the 6D pose of the end-effector~\cite{liang_dreamitate_2025} or the object~\cite{patel_robotic_2025}.
While promising, these approaches require extensive robot-specific data to train a domain-specific model tailored to a particular  embodiment, environment, or task~\cite{du_learning_2023, liang_dreamitate_2025, bharadhwaj_gen2act_2024, ajay_compositional_2023, li_unified_2025, liang_video_2025}.

A key limitation of many video-based manipulation methods is their reliance on embodiment-dependent action generation, which hinders cross-embodiment generalization. 
To address this, object-centric approaches have been proposed. 
For example, a concurrent work~\cite{patel_robotic_2025} extracts 6D poses from the generated video for demonstration-free manipulation, which is object-centric and generalizes across embodiments. 
However, it is model-based and relies on a rigid-body assumption, limiting its applicability to a broader class of objects. 
To achieve greater object generalization, a shift towards model-free representations is essential, which then motivates the adoption of flow-based approaches.

\subsection{Flow-based Manipulation}
\label{sec:flow-based-manipulation}
Flow describes object motion by tracking the displacement of 2D pixels or 3D points between video frames. 
This offers a more generalizable representation of object dynamics compared to 6D pose, as it is inherently model-free and makes no assumptions about object rigidity.
Recent work has shown success in using flow for manipulation~\cite{yuan_general_2024,eisner_flowbot3d_2024,yin_object-centric_2025,bharadhwaj_track2act_2024,xu_flow_2025,ko_learning_2023}.
However, these methods require robot data or task-specific training for either the flow generator or the executor~\cite{yuan_general_2024,yin_object-centric_2025,bharadhwaj_track2act_2024,xu_flow_2025,ko_learning_2023,eisner_flowbot3d_2024,zhang_flowbot_2024}.
To achieve greater generalization for zero-shot manipulation, \citet{chen_vidbot_2025} and \citet{zhi_3dflowaction_2025} train a flow generator on a collection of large-scale human egocentric datasets.
While making a great step towards generalization, we empirically find that the generalizability of this approach~\cite{chen_vidbot_2025} (understanding of in-the-wild object motion) is still not as good as the commonsense motion knowledge from pretrained video models.

\section{\shortname}
\shortname enables robots to autonomously solve a wide variety of manipulation tasks by leveraging pretrained video generation models, thus eliminating the need for demonstrations or task-specific tuning.
Since raw video pixels cannot be directly used by a robot's controller or model-based planner, \shortname handles this challenge by distilling the video's implicit commonsense knowledge of motion into a more actionable, intermediate representation: 3D object flow. 
The proposed pipeline consists of two core components: a \emph{flow generator} (\cref{fig:flow-generator}) that extracts the actionable 3D object flow from the generated video, and a \emph{flow executor} (\cref{fig:flow-executor}) that translates this flow into robot actions. 
The entire pipeline is demonstration-free and embodiment-agnostic, requiring no robot-specific data or training before task execution.

\subsection{Flow Generator} 
The primary objective of the flow generator is to translate a high-level task description into a structured, actionable flow for the robot.  The standard input to the generator is the task description, which involves an initial RGB-D image pair $\{\mathbf{I}, \mathbf{D}\}$ captured from the robot's perspective (with known camera intrinsics) and a natural language instruction, $\mathbf{l}$, describing the desired task. For tasks requiring greater precision, an optional goal image, $\mathbf{I}_g$, can also be provided, either specified by the user~\cite{bharadhwaj_track2act_2024,shi_zeromimic_2025,zhou_dino-wm_2025} or generated by an image editing model with the standard input~\cite{labs_flux1_2025}. Based on the given input, the generator's goal is to produce a 3D object flow across $T$ frames for $M$ object keypoints, $\mathcal{F} \in \mathbb{R}^{T \times M \times 3}$.

The flow generator synthesizes a video using the task description and then distills an \emph{actionable 3D object flow} from the generated video with a pipeline of pre-trained perception modules.
The full process involves five main steps: 
(1) generating the video from the initial image and text prompt,
(2) lifting the 2D video to 3D,
(3) calibrating the estimated depth,
(4) tracking dense per-point 3D motion, and
(5) extracting the final object-centric 3D flow. 

\subsubsection{Video Generation}
Given the initial image $\mathbf{I}$ and language prompt $\mathbf{l}$, a video generation model produces a video $\hat{\mathbf{V}}=\{ \mathbf{\hat{I}}_1, \mathbf{\hat{I}}_2, \ldots, \mathbf{\hat{I}}_T \}$ of $T$ frames, known as image-to-video (I2V) generation. 
If a goal image $\mathbf{I}_g$ is provided, we use first-last-frame-to-video (FLF2V) generation instead.

\subsubsection{Monocular Depth Estimation}
To obtain 3D motion information, we lift the generated 2D video into 3D space. 
We apply a monocular video depth estimation model to  $\hat{\mathbf{V}}$, which processes the video frame-by-frame to yield a sequence of estimated metric depth maps $\mathbf{\hat{D}} = \{\mathbf{\hat{D}}_1, \mathbf{\hat{D}}_2, \ldots, \mathbf{\hat{D}}_T \}$.

\subsubsection{Depth Calibration}
The depth maps $\mathbf{\hat{D}}$ obtained in the last step have a key limitation: the monocular depth estimation process is inherently ill-posed and  often creates metric outputs with systematic scaling errors, especially on generated videos.  This can hinder manipulation tasks that require accurate spatial alignment.
To correct for this, we calibrate the entire estimated depth sequence $\mathbf{\hat{D}}$ by anchoring it to the initial ground-truth depth map. This calibration leverages the observation that estimated depth, while globally inaccurate, is often locally consistent. We compute a scaling factor between the median depth of the first estimated frame $\mathbf{\hat{D}}_1$ and the initial ground-truth depth map $\mathbf{D}$. 
While other methods exist, e.g., fitting an affine transformation~\cite{patel_robotic_2025},  we find this median scaling factor method to be more stable.

\subsubsection{3D Point Tracking}
With the generated video and the calibrated depth, we extract dense per-point 3D motion.
We employ a 3D point tracking model, which takes the camera intrinsics, video $\hat{\mathbf{V}}$, the calibrated depth $\mathbf{\hat{D}}$,  and a set of query points $\mathcal{Q}=\{ \mathbf{q}_1, \ldots, \mathbf{q}_M \}$ evenly sampled on the first frame as input. 
The model outputs a set of 3D trajectories $\mathcal{P} = \{\mathbf{p}_1^t, \ldots, \mathbf{p}_M^t \}_{t=1}^T$, where $\mathbf{p}_i^t$ is the 3D position of the $i$-th query point at timestep $t$.

\subsubsection{Object Grounding}
The dense 3D trajectories $\mathcal{P}$ capture the motion of the entire scene. To derive an actionable plan, we must now ground this motion by isolating only the trajectories belonging to the target objects.
We achieve this by employing a pipeline that combines an open-vocabulary object detector with a video segmentation model,
which produces a sequence of masks, $\mathcal{M} = \{\mathbf{m}_1, \ldots, \mathbf{m}_T \}$, that segment the object across the entire video.
Lastly, by applying these masks, we filter the dense trajectories $\mathcal{P}$ to distill the actionable 3D object flow, $\mathcal{F} = \{ \mathbf{f}_i^t \mid i=1,\ldots,K;\ t=1,\ldots,T \}$. 
This final output represents the $K$ keypoints that remain consistently  tracked on the object's surface.

\begin{figure*}
    \centering
    \includegraphics[width=\textwidth]{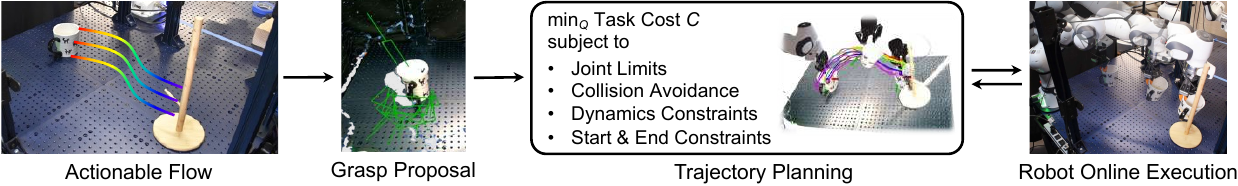}
    \caption{
        \textbf{Flow executor pipeline.} The initial end-effector pose is determined from grasp proposal candidates. Robot trajectories are then planned based on the actionable flow considering costs and constraints, and subsequently tracked by the robots.
    }
    \vspace{-1em}
    \label{fig:flow-executor}
\end{figure*}

\begin{figure}[t!]
    \centering
    \includegraphics[width=\columnwidth]{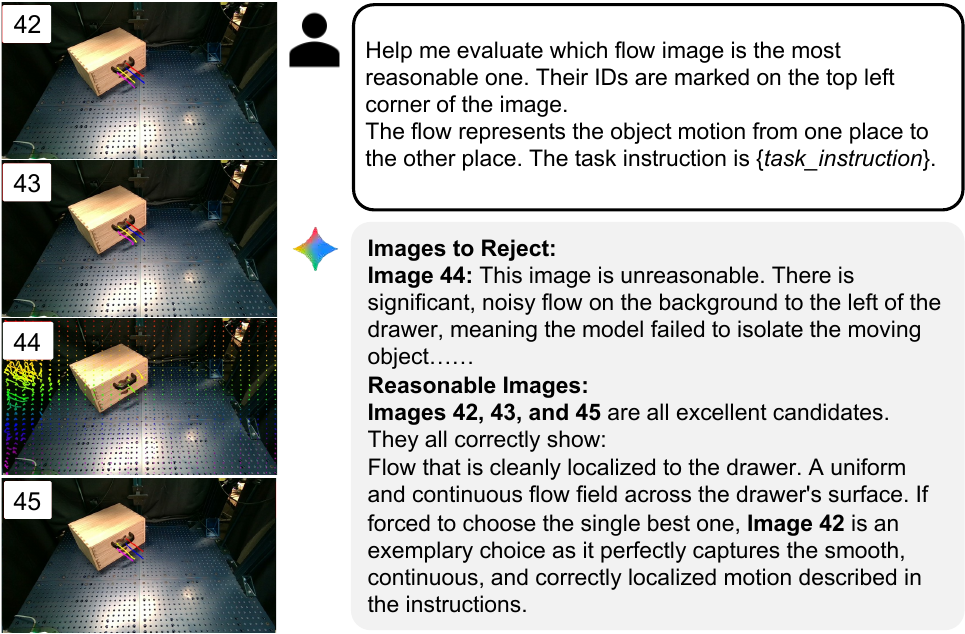}
    \caption{
        \textbf{Rejection sampling for flow generator.}
        We generate multiple video candidates in parallel and create the object flow image for each by back-projecting its object flow, $\mathcal{F}$, onto the initial frame. A VLM (in our case, Google Gemini) evaluates all the flow images to select the most plausible video candidate.
    }
    \vspace{-1em}
    \label{fig:vlm}
\end{figure}

We conclude object grounding with a rejection sampling step to filter out hallucinations, such as generative artifacts and implausible motions, that may be unavoidably introduced by the video generation model (Fig.~\ref{fig:vlm}). Here, we use a VLM to validate and select the most plausible generated flow. Specifically, we generate $N$ video candidates simultaneously and obtain $N$ corresponding object flow images by back-projecting object flow $\mathcal{F}$ to the first frame of each video.
We then mark each flow image using its ID and pass it into a VLM along with its task description to select the most plausible one.
We empirically find that rejecting the flow image is more effective than rejecting the concatenated raw video images~\cite{patel_robotic_2025}, since the flow images,  explicitly visualizing motion, are a more grounded and direct input for VLMs to reason and understand.

\subsection{Flow Executor}
The flow executor is responsible for translating the abstract 3D object flow $\mathcal{F}$ as planned trajectories into a sequence of executable robot actions $\mathcal{A} = \{\mathbf{a}_1, \mathbf{a}_2, \ldots, \mathbf{a}_T\}$ over $T$ timesteps.  The input actionable flow serves as an intermediate representation that describes the desired trajectories of $K$ points on the target object over $T$ timesteps, carrying the high-level task understanding from pre-trained video models.
Here, we present an open-loop planner that can be extended to a closed-loop tracking system by incorporating live object trackers.
The current executor pipeline can handle two main classes of objects: rigid (including articulated objects, treated as part-wise rigid) and deformable.

\subsubsection{Rigid Object Manipulation}
For rigid body manipulation, the 3D flow of the keypoints, $\mathcal{F}$, can be used to estimate the rigid transforms, $(\mathbf{R}, \mathbf{t})$, of the object across frames in a model-free manner. 
In cases where the object is firmly grasped and moves rigidly with the end-effector (\eg, no slippage), a common assumption in prior work~\cite{patel_robotic_2025,bharadhwaj_track2act_2024,yin_object-centric_2025, huang_imagination_2024}, the end-effector pose can be calculated from the object pose.
The object-specific firm grasps are selected from the object point cloud using a grasp proposal model, as shown in \cref{fig:flow-executor}.
At each timestep $t$, we find the rigid transformation $(\mathbf{R}^t, \mathbf{t}^t)$ that aligns the initial keypoints $\{\mathbf{f}_i^1\}_{i=1}^K$ to the current one $\{\mathbf{f}_i^t\}_{i=1}^K$.
This is solved using the Kabsch algorithm~\cite{kabsch_solution_1976}, which finds the optimal rotation $\mathbf{R}^t$ that minimizes the sum of squared errors:
\begin{equation}
\mathbf{R}^t = \argmin_{\mathbf{R} \in SO(3)} \sum_{i=1}^{K} \| \mathbf{R}(\mathbf{f}_i^1 - \mathbf{c}^1) - (\mathbf{f}_i^t - \mathbf{c}^t) \|^2,
\end{equation}
where $\mathbf{c}^1$ and $\mathbf{c}^t$ are the masked point cloud centroids at the first and current timesteps, respectively. 
This optimization can be solved efficiently using Singular Value Decomposition (SVD). 
Once the rotation is found, the translation is computed as $\mathbf{t}^t = \mathbf{c}^t - \mathbf{R}^t \mathbf{c}^1$. 
The object pose at timestep $t$ can be represented as a homogeneous transformation matrix ${\mathbf{T}_{obj}^t \in SE(3)}$, constructed from $\mathbf{R}^t$ and $\mathbf{t}^t$.
The resulting sequence of 6D object poses is converted into an end-effector trajectory by applying a grasp transformation, $\mathbf{T}_{grasp}$, obtained from a grasping network~\cite{murali_graspgen_2025}. 
The target end-effector pose at each timestep is then:
\begin{equation}
    \mathbf{T}_{ee}^t = \mathbf{T}_{obj}^t \cdot \mathbf{T}_{grasp}.
\end{equation}
This Cartesian pose is converted to joint commands via trajectory optimization for execution by the robot's controller.
  
\subsubsection{Deformable Object Manipulation}
Unlike rigid objects, deformable objects have complex dynamics that cannot be described by a simple rigid transformation.
\shortname can be naturally extended to handle deformable objects, with the 3D object flow $\mathcal{F}$ serving as a dense tracking objective for model-based planning.
Specifically, we employ a particle-based dynamics model $f_{\theta}$ to predict the object's future state, where $\theta$ represents the learnable parameters of the model.
The state of the object at time $t$ is represented by a set of $N_p$ particles $\mathcal{S}_t = \{\mathbf{s}_i^t\}_{i=1}^{N_p}$. 
The dynamics model predicts the next state based on the current state and a robot action $\mathbf{a}_t$: $\mathcal{S}_{t+1} = f_{\theta}(\mathcal{S}_t, \mathbf{a}_t)$.

Conventional methods for deformable manipulation often define a cost function using a correspondence-free metric, like the Chamfer distance, to a single goal state $\mathcal{S}_{goal}$~\cite{jiang_phystwin_2025, zhang_particle-grid_2025,zhang_adaptigraph_2024,wang_dynamic-resolution_2023, huang_particleformer_2025}.
Our actionable 3D object flow $\mathcal{F} = \{\mathcal{F}^t\}_{t=1}^T$, where $\mathcal{F}^t = \{\mathbf{f}_i^t\}_{i=1}^{N_p}
$, allows us to define a cost function based on the sum of squared Euclidean distances, leveraging the explicit point-wise correspondences from the flow:
\begin{equation}
    C(\mathcal{S}_t, \mathcal{F}^t) = \sum_{i=1}^{N_p} \| \mathbf{s}_i^t - \mathbf{f}_i^t \|^2.
    \label{eq:deformable_cost}
\end{equation}
This formulation has two potential advantages. First, using point correspondences may create a better-conditioned optimization landscape, as correspondence-free metrics can be susceptible to local minima. Second, tracking a dense flow provides intermediate targets along a desired motion path, rather than relying on only a final goal configuration.

We then frame the control problem as a Model Predictive Control (MPC) task. At each timestep $t$, we solve for an optimal sequence of actions $\mathbf{A}^*_t = \{\mathbf{a}^*_t, \ldots, \mathbf{a}^*_{t+H-1}\}$ over a planning horizon $H$ by minimizing the cumulative cost:
\begin{equation}
    \mathbf{A}^*_t = \argmin_{\mathbf{A}_t} \sum_{j=t}^{j+H-1} C(\mathcal{S}_j, \mathcal{F}^j),
\end{equation}
subject to the \textit{dynamics constraints} $\mathcal{S}_{j+1} = f_{\theta}(\mathcal{S}_j, \mathbf{a}_j)$. We then execute the first action $\mathbf{a}^*_t$ and repeat the optimization at the next timestep.

\textbf{Trajectory optimization.} To enable smooth and collision-free motion, we additionally incorporate trajectory optimization to refine the sequence of actions.
We formulate the trajectory generation as a non-linear least-squares problem. 
The goal is to find an optimal sequence of joint configurations $Q = \{q_0, q_1, \dots, q_{T-1}\}$ that minimizes a sum-of-squares objective function. 
The trajectory is initialized by linearly interpolating between start and end configurations, $q_{\text{start,IK}}$ and $q_{\text{end,IK}}$, which are pre-calculated using an IK solver using the end-effector pose.   
The optimal trajectory $Q^*$ is found by solving the constrained non-linear optimization problem:
\begin{equation}
\begin{aligned}
\min_{Q} \quad & w_s \mathcal{C}_{\text{smooth}} + w_r \mathcal{C}_{\text{rest}}, \text{\quad subject to} \\
\quad & q_0 = q_{\text{start,IK}} \quad \text{and} \quad q_{T-1} = q_{\text{end,IK}}, \\
& q_{\min} \le q_t \le q_{\max}, \quad \forall t \in \{0, \dots, T-1\}, \\
& d_s(q_t, q_{t+1}, O_j) \ge \epsilon_{\text{safe}}, \quad \forall t, \forall O_j \in \text{Obstacles}.
\end{aligned}
\end{equation}

In this formulation, the objective function seeks to minimize a weighted sum of the motion smoothness cost ($\mathcal{C}_{\text{smooth}}$) and the rest pose regularization cost ($\mathcal{C}_{\text{rest}}$). 
Constraints in the optimization include:
(1) \textit{start and end constraints}, meaning the trajectory's start and end configurations ($q_0$ and $q_{T-1}$) must exactly match the predefined goals ($q_{\text{start, IK}}$ and $q_{\text{end, IK}}$);
(2) \textit{collision avoidance}, by enforcing the signed distance, $d_s$, between the robot and any obstacle to remain greater than a safety margin, $\epsilon_{\text{safe}}$, at all times;
(3) \textit{joint limits}, ensuring the robot's physical joint position and velocity limits throughout the entire motion.
We treat these constraints as cost terms and use the Levenberg-Marquardt solver to solve the non-linear least-squares problem.

\begin{figure*}[t!]
    \centering
    \includegraphics[width=\linewidth]{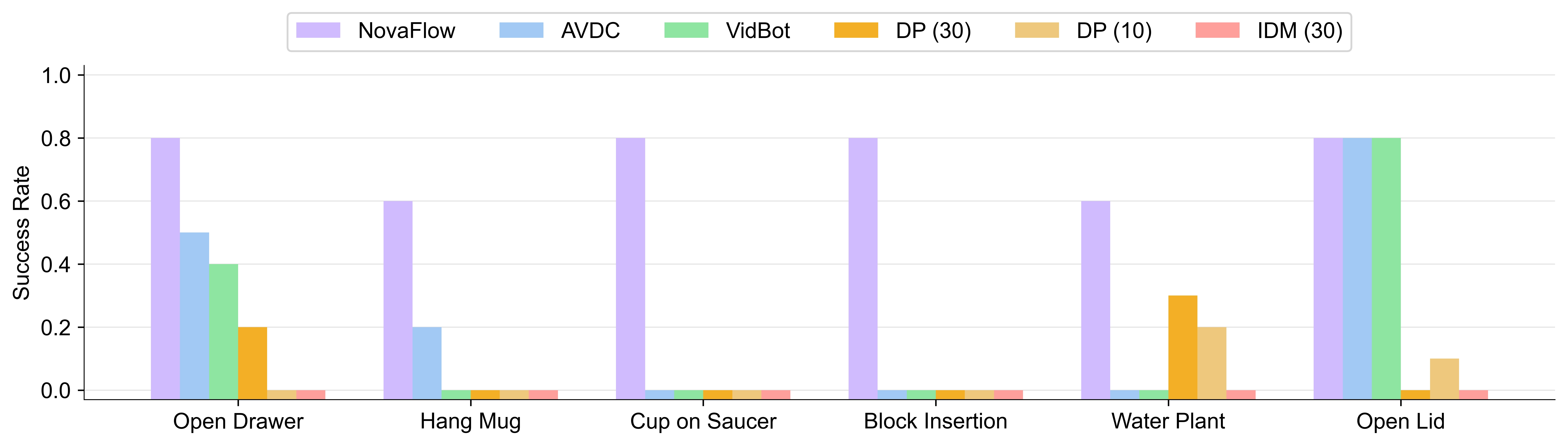}
    \caption{
        \textbf{Experiment results.}
        We compare against Diffusion Policy (DP)~\cite{chi_diffusion_2023} trained using 10 and 30 demonstrations, inverse dynamics model (IDM) from UniPi~\cite{du_learning_2023}, AVDC~\cite{ko_learning_2023}, and VidBot~\cite{chen_vidbot_2025} in real-world tabletop manipulation tasks.
    }
    \label{fig:experiment_results}
    \vspace{-1em}
\end{figure*}

\section{Experiments}
We aim to demonstrate the generalizability of \shortname across different object types and embodiments and to show the importance of each component in our framework. 
We evaluate the framework's ability to execute a broad range of manipulation tasks involving rigid, articulated, and deformable objects across embodiments without requiring task-specific demonstrations or additional fine-tuning.

\subsection{Implementation Details}
We implement \shortname with modular, swappable components. 
For video generation, we use the open-source model Wan~\cite{wan_wan_2025}, which produces 41 frames per task (16\,FPS, 1280$\times$720).
We estimate depth with MegaSaM~\cite{li_megasam_2025} using calibrated intrinsics, track 3D points with TAPIP3D~\cite{zhang_tapip3d_2025}, and ground objects via Grounded-SAM2~\cite{ren_grounded_2024} (Grounding DINO~\cite{liu_grounding_2025} + SAM2~\cite{ravi_sam_2024}). 
For flow executor, we use GraspGen~\cite{murali_graspgen_2025} to generate grasp proposals.
We use a trained PhysTwin~\cite{jiang_phystwin_2025} model to predict particle dynamics for deformable objects. 
All modules are drop-in replaceable with newer models, improving speed and robustness, which is another benefit for our modular framework.

\subsection{Real-World Experiments and Evaluation Tasks}
We evaluate \shortname on a Franka arm with a Robotiq-85 gripper for table-top manipulation and a Spot quadruped for mobile robot manipulation.
For rigid and articulated objects, we use a single RealSense D455 depth camera as input. 
For deformable objects, we use three synchronized cameras (as required by PhysTwin~\cite{jiang_phystwin_2025}), though a single-view setup is also possible~\cite{huang_particleformer_2025}.

\begin{figure*}[t!]
    \centering
    \includegraphics[width=\linewidth]{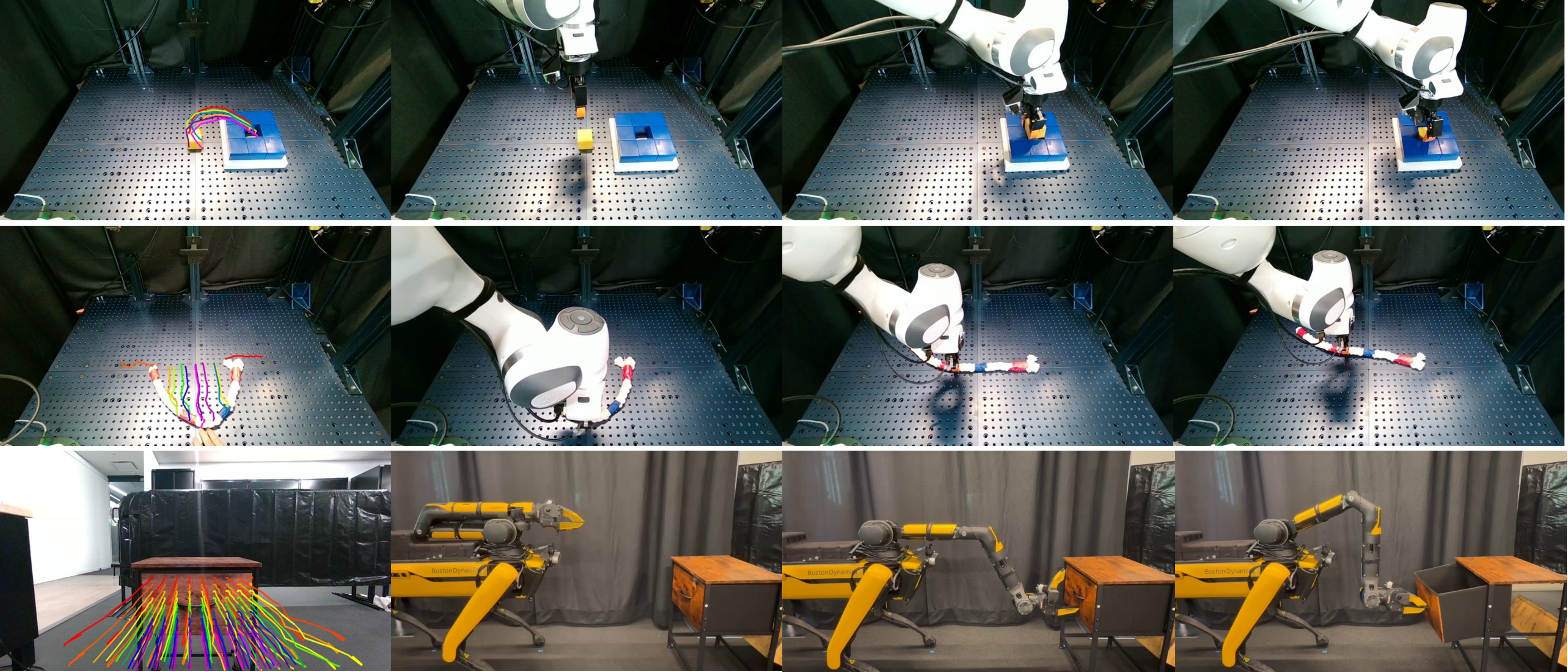}
    \caption{
        \textbf{Real-world manipulation experiments.}
        \shortname is versatile and supports cross-embodiment manipulation, which we use to manipulate rigid, deformable, and articulated objects using tabletop and mobile manipulator.
    }
    \label{fig:real_world_experiments}
    \vspace{-1em}
\end{figure*}

We categorize our tasks by the object type involved as rigid (\textbf{R}), articulated (\textbf{A}), and deformable (\textbf{D}):
\begin{itemize}
  \item \textbf{Hanging a mug (R)}: hang a mug on a wooden rack, requiring accurate relative pose placement for the handle to pass through the wooden stick on the rack.
  \item \textbf{Inserting a block (R)}: insert a yellow block into a hole in a board, a task similar to peg-in-hole that requires accurate insertion skills.
  \item \textbf{Placing a cup on a saucer (R)}: place a cup on a saucer, a task demanding accurate placement skills.
  \item \textbf{Watering a plant (R)}: pour water from a green cup into a plant pot, requiring language understanding and manipulation skills.
  \item \textbf{Opening a drawer (A)}: open a drawer, requiring a precise understanding of its articulation.
  \item \textbf{Straightening a rope (D)}: straighten a curved rope, which requires understanding the dynamics of a deformable object.
\end{itemize}
During evaluation, we randomize the object placement after each trial.
We report the quantitative and qualitative results in \cref{fig:experiment_results} and \cref{fig:real_world_experiments}.

\subsection{Comparison with Baselines}
We compare \shortname against two groups of baselines. (+) denotes methods requiring external training data, while (*) denotes baselines adapted to fit our pipeline.

\textbf{Demo-free, zero-shot baselines (similar to ours):}
\begin{itemize}
  \item \textbf{AVDC~\cite{ko_learning_2023} (*):} Extracts object-centric motion using optical flow. We adapt it to our pipeline by applying it directly to generated videos.
  \item \textbf{VidBot~\cite{chen_vidbot_2025}:} Learns flow from large-scale human interaction datasets to model affordances.
\end{itemize}

\textbf{Data-dependent baselines (require demonstrations):}
\begin{itemize}
  \item \textbf{Diffusion Policy (DP)~\cite{chi_diffusion_2023} (+):} Diffusion policy serves as an imitation policy baseline trained under very few demos for a single task. We train with 10 and 30 demonstrations per task, using the same single-view camera RGB input as our approach.
  \item \textbf{Inverse Dynamics Model (IDM)~\cite{du_learning_2023} (+):} IDM was originally designed to train together with a fine-tuned video generation model using in-domain demonstrations. Since video fine-tuning is outside our scope, we trained the IDM model with the 30 demonstrations previously used in DP training to convert generated robot task-solving videos (from Wan2.1) into robot actions.
\end{itemize}

\shortname achieves the highest success rates across tasks among zero-shot methods and also surpasses data-dependent baselines trained with 10–30 demonstrations, as shown in \cref{fig:experiment_results} (with 10 trials for each task).
\textbf{AVDC(*)} performs competitively on affordance-like tasks but struggles with precise, long-horizon placements. In our setup, it distills motion from 2D optical flow, lacking 3D awareness and long-term coherence under occlusion. 
These limitations, as also noted in the AVDC paper, cause the method to struggle with tasks requiring accurate placement and rotation-heavy motions. 
\textbf{VidBot} excels on affordance-centric, articulated interactions (e.g., “open drawer”) but fails when tasks require object–object relations and precise relative pose placement. 
This matches our diagnosis that its training emphasizes object–affordance understanding rather than modeling multi-object constraints. 
For \textbf{DP(+)}, despite per-task training (an easier setting that bypasses language understanding), it still shows poor generalization from a few examples, especially because our evaluations are randomly sampled and not drawn from the training distribution. 
The main issue for \textbf{IDM(+)} is the domain shift between its training and test data. 
The inverse dynamics model learns from real-world robot demos, yet it must interpret generated videos whose motion is not always kinematically perfect or consistent. 
Consequently, the generated videos are out-of-distribution, causing the model to fail even if the video's high-level action seems semantically reasonable.

Overall, methods that (i) lack an actionable \emph{3D} representation (AVDC, VidBot) or (ii) rely on small, task-specific robot datasets (DP, IDM) fail to provide zero-shot autonomous task-solving. Distilling a dense, actionable 3D object flow and decoupling understanding from control is critical for zero-shot generalization.

\begin{figure}[t!]
    \centering
    \includegraphics[width=\linewidth]{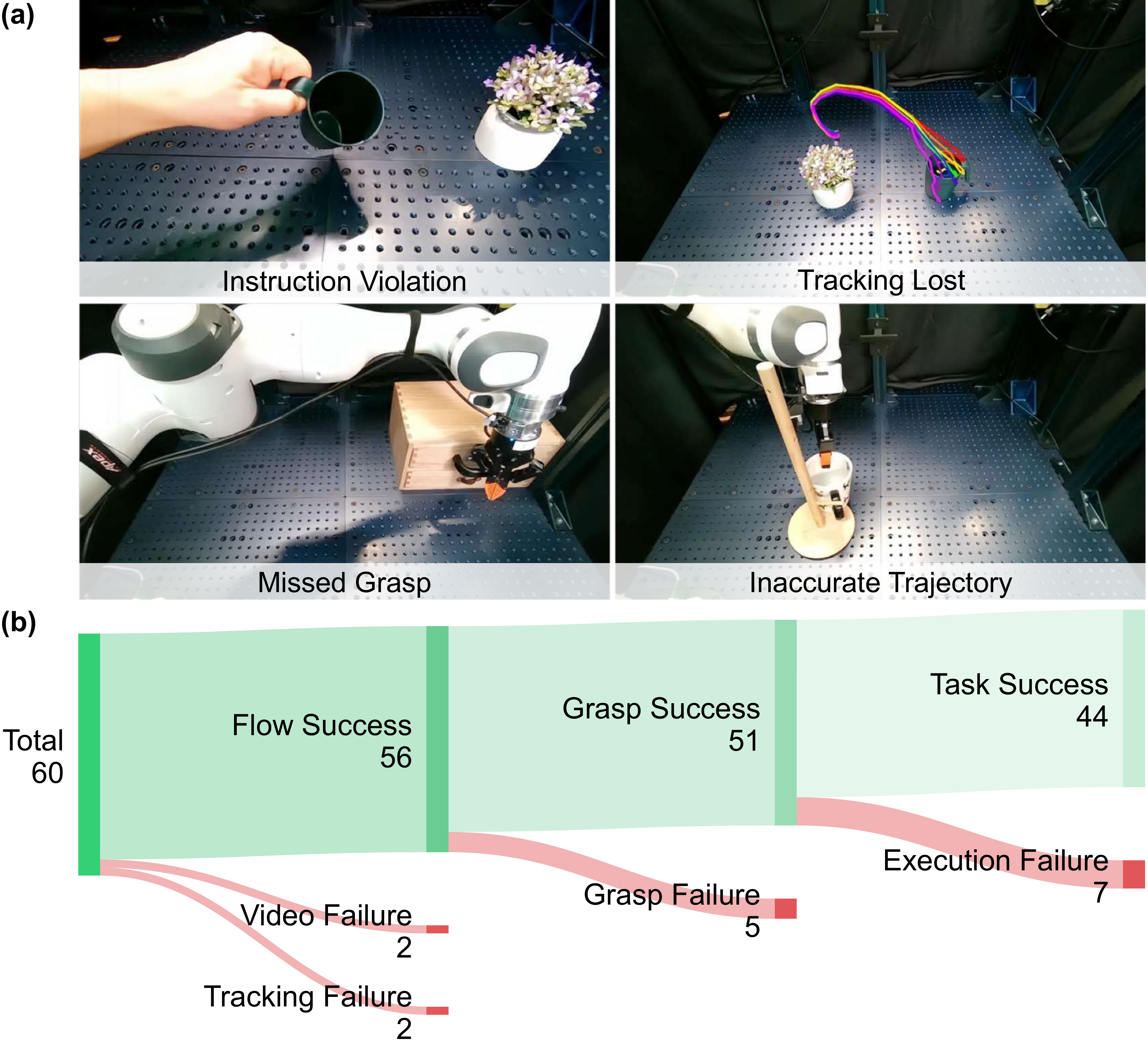}
    \caption{
        \textbf{Failure analysis.}
       (a) Examples of video, tracking, grasp, and execution failures. (b) Failure cause distribution.
    }
    \label{fig:failure_analysis}
    \vspace{-5pt}
\end{figure}
\subsection{Ablation Studies and Failure Analysis}
Here we discuss some of the design choices in \shortname's submodules. 
For video generation, we compare the current Wan 2.1 model to the closed-source model Veo~\cite{wiedemer_video_2025}, which produces 8\,s clips (24\, FPS). 
Prompt extension is utilized for better controllability.
For precise placement tasks (e.g., mug on rack and block insertion), we optionally condition on a goal image (FLF2V) instead of I2V.

We analyze the failure cases of \shortname in \cref{fig:failure_analysis}, identifying four primary failure modes. 
\textbf{Video failure} occurs when the generative model produces content that is not physically plausible, lacks 3D consistency, or violates the user's instructions; our rejection sampling with a VLM mitigates but does not eliminate this.
\textbf{Tracking failure} results from inaccuracy in 3D point tracking, often caused by textureless surfaces, heavy occlusions, or accumulated inconsistencies inherited from the video model. 
\textbf{Grasp failure} happens when the robot fails to secure the object correctly (e.g., bad approach, missed grasp, and slip).
Finally, \textbf{execution failure} encompasses errors during trajectory execution, such as collisions, joint limits, or an inability to follow the planned path accurately. 
Our analysis reveals that most failures occur in the \emph{last mile}: grasp and execution are the most frequent, suggesting that while the upstream flow estimation is relatively robust, physical interaction remains the bottleneck. 
This is similar to the sim-to-real gap in simulation-based training. 
To address these limitations, future work could focus on integrating a closed-loop feedback system to enable dynamic replanning and refine the generated flow in response to observations.
\begin{table}[t!]
\centering
\caption{Effect of goal image on block insertion task.}
\label{tab:goal_image_effect}
\autosizeTable{
\begin{tabular}{lccc} %
\hline
\textbf{Condition} & \textbf{Video Success} & \textbf{Task Success} & \textbf{Time (s)} \\
\hline
w/ Goal Image (Wan2.1) & 46\% & 80\% & 612 \\
w/o Goal Image (Wan2.1) & 15\% & 40\% & 612 \\
w/o Goal Image (Veo) &   75\% & 80\% & 20 \\
\hline
\end{tabular}
}
\vspace{-5pt}
\end{table}

We investigate the effect of a goal image on the block insertion task requiring millimeter-level precision (\cref{tab:goal_image_effect}). 
We use two metrics: Video Success Rate, the percentage of generated videos with a valid actionable flow, and Task Success Rate, the execution success of a flow selected by a VLM after rejection sampling. 
For each trial, we synthesize eight videos, from which the VLM selects the best for execution. 
Our results show that omitting the goal image significantly impairs the performance of the open-source Wan2.1 model. 
While VLM rejection sampling improves the final success rate, the drop remains substantial. 
In contrast, the closed-source Veo model proves more robust, outperforming Wan2.1 and achieving a high task success rate even without a goal image.

\subsection{Runtime Analysis}
\label{sec:runtime}
We deploy \shortname on a single NVIDIA H100 GPU, and a complete flow generation takes around 2 minutes end-to-end (Veo).  
We report per-module timings in \cref{tab:runtime} to guide replacements and optimization. 
The dominant time-consuming modules are the video generation and 3D lifting modules. 
For video generation, closed-source models are usually much faster in time but more expensive in cost.
\begin{table}[t!]
\centering
\caption{Runtime analysis. Time is measured in seconds.}
\label{tab:runtime}
\autosizeTable{
\begin{tabular}{lccccc}
\hline
  & MegaSaM & TAPIP3D & SAM2 & \textbf{Total (Veo)} & Total (Wan) \\
\hline
\textbf{Time} & 100 & 5 & 8 & \textbf{133} & 725 \\
\hline
\end{tabular}
}
\vspace{-2em}
\end{table}

\section{Conclusion}
We introduced \shortname, a demonstration-free framework for autonomous manipulation that translates natural language commands into robot actions by leveraging the commonsense knowledge embedded in large-scale video generation models. 
Our key insight is to distill generated task-solving videos into an actionable 3D object flow, an intermediate representation that decouples high-level task understanding from low-level robot control.
This modular design enables \shortname to handle rigid, articulated, and deformable objects across different robot embodiments without requiring any task-specific training or demonstrations.
Our real-world experiments show that \shortname not only outperforms other zero-shot methods but also surpasses imitation learning policies trained on dozens of demonstrations.

Despite its success, our failure analysis reveals that the primary bottleneck is the physical execution phase, particularly in grasping and handling unexpected dynamics. 
This highlights a gap between the open-loop plan generated from video and the complexity of real-world interaction.
A promising direction for future work is to develop a closed-loop system where real-time feedback from the environment is used to refine or replan the generated flow, making the system more adaptive and robust to unforeseen challenges.

\bibliographystyle{custom-ieee}
\bibliography{custom, local}

% Generated by IEEEtranN.bst, version: 1.14 (2015/08/26)
\begin{thebibliography}{62}
\providecommand{\natexlab}[1]{#1}
\providecommand{\url}[1]{#1}
\csname url@samestyle\endcsname
\providecommand{\newblock}{\relax}
\providecommand{\bibinfo}[2]{#2}
\providecommand{\BIBentrySTDinterwordspacing}{\spaceskip=0pt\relax}
\providecommand{\BIBentryALTinterwordstretchfactor}{4}
\providecommand{\BIBentryALTinterwordspacing}{\spaceskip=\fontdimen2\font plus
\BIBentryALTinterwordstretchfactor\fontdimen3\font minus \fontdimen4\font\relax}
\providecommand{\BIBforeignlanguage}[2]{{%
\expandafter\ifx\csname l@#1\endcsname\relax
\typeout{** WARNING: IEEEtranN.bst: No hyphenation pattern has been}%
\typeout{** loaded for the language `#1'. Using the pattern for}%
\typeout{** the default language instead.}%
\else
\language=\csname l@#1\endcsname
\fi
#2}}
\providecommand{\BIBdecl}{\relax}
\BIBdecl

\bibitem[Kim et~al.(2025{\natexlab{a}})Kim, Pertsch, Karamcheti, Xiao, Balakrishna, Nair, Rafailov, Foster, Sanketi, Vuong, Kollar, Burchfiel, Tedrake, Sadigh, Levine, Liang, and Finn]{kim_openvla_2025}
M.~J. Kim, K.~Pertsch, S.~Karamcheti, T.~Xiao, A.~Balakrishna, S.~Nair, R.~Rafailov, E.~P. Foster, P.~R. Sanketi \emph{et~al.}, ``\BIBforeignlanguage{en}{{OpenVLA}: {An} {Open}-{Source} {Vision}-{Language}-{Action} {Model}},'' in \emph{\BIBforeignlanguage{en}{Conference on {Robot} {Learning} ({CoRL})}}, Jan. 2025.

\bibitem[Brohan et~al.(2022)Brohan, Brown, Carbajal, Chebotar, Dabis, Finn, Gopalakrishnan, Hausman, Herzog, Hsu, Ibarz, Ichter, Irpan, Jackson, Jesmonth, Joshi, Julian, Kalashnikov, Kuang, Leal, Lee, Levine, Lu, Malla, Manjunath, Mordatch, Nachum, Parada, Peralta, Perez, Pertsch, Quiambao, Rao, Ryoo, Salazar, Sanketi, Sayed, Singh, Sontakke, Stone, Tan, Tran, Vanhoucke, Vega, Vuong, Xia, Xiao, Xu, Xu, Yu, and Zitkovich]{brohan_rt-1_2022}
A.~Brohan, N.~Brown, J.~Carbajal, Y.~Chebotar, J.~Dabis, C.~Finn, K.~Gopalakrishnan, K.~Hausman, A.~Herzog \emph{et~al.}, ``{RT}-1: {Robotics} {Transformer} for {Real}-{World} {Control} at {Scale},'' Dec. 2022, arXiv:2212.06817 [cs].

\bibitem[Ghosh et~al.(2024)Ghosh, Walke, Pertsch, Black, Mees, Dasari, Hejna, Kreiman, Xu, Luo, Tan, Chen, Vuong, Xiao, Sanketi, Sadigh, Finn, and Levine]{ghosh_octo_2024}
D.~Ghosh, H.~R. Walke, K.~Pertsch, K.~Black, O.~Mees, S.~Dasari, J.~Hejna, T.~Kreiman, C.~Xu \emph{et~al.}, ``Octo: {An} {Open}-{Source} {Generalist} {Robot} {Policy},'' in \emph{Robotics: {Science} and {Systems} ({RSS})}, vol.~20, Jul. 2024.

\bibitem[Team et~al.(2025)Team, Barreiros, Beaulieu, Bhat, Cory, Cousineau, Dai, Fang, Hashimoto, Irshad, Itkina, Kuppuswamy, Lee, Liu, McConachie, McMahon, Nishimura, Phillips-Grafflin, Richter, Shah, Srinivasan, Wulfe, Xu, Zhang, Alspach, Angeles, Arora, Guizilini, Castro, Chen, Chu, Creasey, Curtis, Denitto, Dixon, Dusel, Ferreira, Goncalves, Gould, Guoy, Gupta, Han, Hatch, Hathaway, Henry, Hochsztein, Horgan, Iwase, Jackson, Karamcheti, Keh, Masterjohn, Mercat, Miller, Mitiguy, Nguyen, Nimmer, Noguchi, Ong, Onol, Pfannenstiehl, Poyner, Rocha, Richardson, Rodriguez, Seale, Sherman, Smith-Jones, Tago, Tokmakov, Tran, Hoorick, Vasiljevic, Zakharov, Zolotas, Ambrus, Fetzer-Borelli, Burchfiel, Kress-Gazit, Feng, Ford, and Tedrake]{team_careful_2025}
T.~L. Team, J.~Barreiros, A.~Beaulieu, A.~Bhat, R.~Cory, E.~Cousineau, H.~Dai, C.-H. Fang, K.~Hashimoto \emph{et~al.}, ``A {Careful} {Examination} of {Large} {Behavior} {Models} for {Multitask} {Dexterous} {Manipulation},'' Jul. 2025, arXiv:2507.05331.

\bibitem[Qwen et~al.(2025)Qwen, Yang, Yang, Zhang, Hui, Zheng, Yu, Li, Liu, Huang, Wei, Lin, Yang, Tu, Zhang, Yang, Yang, Zhou, Lin, Dang, Lu, Bao, Yang, Yu, Li, Xue, Zhang, Zhu, Men, Lin, Li, Tang, Xia, Ren, Ren, Fan, Su, Zhang, Wan, Liu, Cui, Zhang, and Qiu]{qwen_qwen25_2025}
Qwen, A.~Yang, B.~Yang, B.~Zhang, B.~Hui, B.~Zheng, B.~Yu, C.~Li, D.~Liu \emph{et~al.}, ``Qwen2.5 {Technical} {Report},'' Jan. 2025, arXiv:2412.15115 [cs].

\bibitem[DeepSeek-AI et~al.(2025)DeepSeek-AI, Guo, Yang, Zhang, Song, Zhang, Xu, Zhu, Ma, Wang, Bi, Zhang, Yu, Wu, Wu, Gou, Shao, Li, Gao, Liu, Xue, Wang, Wu, Feng, Lu, Zhao, Deng, Zhang, Ruan, Dai, Chen, Ji, Li, Lin, Dai, Luo, Hao, Chen, Li, Zhang, Bao, Xu, Wang, Ding, Xin, Gao, Qu, Li, Guo, Li, Wang, Chen, Yuan, Qiu, Li, Cai, Ni, Liang, Chen, Dong, Hu, Gao, Guan, Huang, Yu, Wang, Zhang, Zhao, Wang, Zhang, Xu, Xia, Zhang, Zhang, Tang, Li, Wang, Li, Tian, Huang, Zhang, Wang, Chen, Du, Ge, Zhang, Pan, Wang, Chen, Jin, Chen, Lu, Zhou, Chen, Ye, Wang, Yu, Zhou, Pan, Li, Zhou, Wu, Ye, Yun, Pei, Sun, Wang, Zeng, Zhao, Liu, Liang, Gao, Yu, Zhang, Xiao, An, Liu, Wang, Chen, Nie, Cheng, Liu, Xie, Liu, Yang, Li, Su, Lin, Li, Jin, Shen, Chen, Sun, Wang, Song, Zhou, Wang, Shan, Li, Wang, Wei, Zhang, Xu, Li, Zhao, Sun, Wang, Yu, Zhang, Shi, Xiong, He, Piao, Wang, Tan, Ma, Liu, Guo, Ou, Wang, Gong, Zou, He, Xiong, Luo, You, Liu, Zhou, Zhu, Xu, Huang, Li, Zheng, Zhu, Ma, Tang, Zha, Yan, Ren, Ren, Sha, Fu, Xu, Xie, Zhang, Hao, Ma, Yan, Wu, Gu, Zhu, Liu, Li, Xie, Song, Pan, Huang, Xu, Zhang, and Zhang]{deepseek-ai_deepseek-r1_2025}
DeepSeek-AI, D.~Guo, D.~Yang, H.~Zhang, J.~Song, R.~Zhang, R.~Xu, Q.~Zhu, S.~Ma \emph{et~al.}, ``{DeepSeek}-{R1}: {Incentivizing} {Reasoning} {Capability} in {LLMs} via {Reinforcement} {Learning},'' Jan. 2025, arXiv:2501.12948 [cs].

\bibitem[Touvron et~al.(2023)Touvron, Lavril, Izacard, Martinet, Lachaux, Lacroix, Rozière, Goyal, Hambro, Azhar, Rodriguez, Joulin, Grave, and Lample]{touvron_llama_2023}
H.~Touvron, T.~Lavril, G.~Izacard, X.~Martinet, M.-A. Lachaux, T.~Lacroix, B.~Rozière, N.~Goyal, E.~Hambro \emph{et~al.}, ``{LLaMA}: {Open} and {Efficient} {Foundation} {Language} {Models},'' Feb. 2023, arXiv:2302.13971 [cs].

\bibitem[Liu et~al.(2023)Liu, Li, Wu, and Lee]{liu_visual_2023}
H.~Liu, C.~Li, Q.~Wu, and Y.~J. Lee, ``\BIBforeignlanguage{en}{Visual {Instruction} {Tuning}},'' in \emph{\BIBforeignlanguage{en}{Advances in {Neural} {Information} {Processing} {Systems} ({NeurIPS})}}, vol.~36, Dec. 2023.

\bibitem[OpenAI et~al.(2024)OpenAI, Hurst, Lerer, Goucher, Perelman, Ramesh, Clark, Ostrow, Welihinda, Hayes, Radford, Madry, Baker-Whitcomb, Beutel, Borzunov, Carney, Chow, Kirillov, Nichol, Paino, Renzin, Passos, Kirillov, Christakis, Conneau, Kamali, Jabri, Moyer, Tam, Crookes, Tootoochian, Tootoonchian, Kumar, Vallone, Karpathy, Braunstein, Cann, Codispoti, Galu, Kondrich, Tulloch, Mishchenko, Baek, Jiang, Pelisse, Woodford, Gosalia, Dhar, Pantuliano, Nayak, Oliver, Zoph, Ghorbani, Leimberger, Rossen, Sokolowsky, Wang, Zweig, Hoover, Samic, McGrew, Spero, Giertler, Cheng, Lightcap, Walkin, Quinn, Guarraci, Hsu, Kellogg, Eastman, Lugaresi, Wainwright, Bassin, Hudson, Chu, Nelson, Li, Shern, Conger, Barette, Voss, Ding, Lu, Zhang, Beaumont, Hallacy, Koch, Gibson, Kim, Choi, McLeavey, Hesse, Fischer, Winter, Czarnecki, Jarvis, Wei, Koumouzelis, Sherburn, Kappler, Levin, Levy, Carr, Farhi, Mely, Robinson, Sasaki, Jin, Valladares, Tsipras, Li, Nguyen, Findlay, Oiwoh, Wong, Asdar, Proehl, Yang, Antonow, Kramer, Peterson, Sigler, Wallace, Brevdo, Mays, Khorasani, Such, Raso, Zhang, Lohmann, Sulit, Goh, Oden, Salmon, Starace, Brockman, Salman, Bao, Hu, Wong, Wang, Schmidt, Whitney, Jun, Kirchner, Pinto, Ren, Chang, Chung, Kivlichan, O'Connell, O'Connell, Osband, Silber, Sohl, Okuyucu, Lan, Kostrikov, Sutskever, Kanitscheider, Gulrajani, Coxon, Menick, Pachocki, Aung, Betker, Crooks, Lennon, Kiros, Leike, Park, Kwon, Phang, Teplitz, Wei, Wolfe, Chen, Harris, Varavva, Lee, Shieh, Lin, Yu, Weng, Tang, Yu, Jang, Candela, Beutler, Landers, Parish, Heidecke, Schulman, Lachman, McKay, Uesato, Ward, Kim, Huizinga, Sitkin, Kraaijeveld, Gross, Kaplan, Snyder, Achiam, Jiao, Lee, Zhuang, Harriman, Fricke, Hayashi, Singhal, Shi, Karthik, Wood, Rimbach, Hsu, Nguyen, Gu-Lemberg, Button, Liu, Howe, Muthukumar, Luther, Ahmad, Kai, Itow, Workman, Pathak, Chen, Jing, Guy, Fedus, Zhou, Mamitsuka, Weng, McCallum, Held, Ouyang, Feuvrier, Zhang, Kondraciuk, Kaiser, Hewitt, Metz, Doshi, Aflak, Simens, Boyd, Thompson, Dukhan, Chen, Gray, Hudnall, Zhang, Aljubeh, Litwin, Zeng, Johnson, Shetty, Gupta, Shah, Yatbaz, Yang, Zhong, Glaese, Chen, Janner, Lampe, Petrov, Wu, Wang, Fradin, Pokrass, Castro, Castro, Pavlov, Brundage, Wang, Khan, Murati, Bavarian, Lin, Yesildal, Soto, Gimelshein, Cone, Staudacher, Summers, LaFontaine, Chowdhury, Ryder, Stathas, Turley, Tezak, Felix, Kudige, Keskar, Deutsch, Bundick, Puckett, Nachum, Okelola, Boiko, Murk, Jaffe, Watkins, Godement, Campbell-Moore, Chao, McMillan, Belov, Su, Bak, Bakkum, Deng, Dolan, Hoeschele, Welinder, Tillet, Pronin, Tillet, Dhariwal, Yuan, Dias, Lim, Arora, Troll, Lin, Lopes, Puri, Miyara, Leike, Gaubert, Zamani, Wang, Donnelly, Honsby, Smith, Sahai, Ramchandani, Huet, Carmichael, Zellers, Chen, Chen, Nigmatullin, Cheu, Jain, Altman, Schoenholz, Toizer, Miserendino, Agarwal, Culver, Ethersmith, Gray, Grove, Metzger, Hermani, Jain, Zhao, Wu, Jomoto, Wu, Shuaiqi, Xia, Phene, Papay, Narayanan, Coffey, Lee, Hall, Balaji, Broda, Stramer, Xu, Gogineni, Christianson, Sanders, Patwardhan, Cunninghman, Degry, Dimson, Raoux, Shadwell, Zheng, Underwood, Markov, Sherbakov, Rubin, Stasi, Kaftan, Heywood, Peterson, Walters, Eloundou, Qi, Moeller, Monaco, Kuo, Fomenko, Chang, Zheng, Zhou, Manassra, Sheu, Zaremba, Patil, Qian, Kim, Cheng, Zhang, He, Zhang, Jin, Dai, and Malkov]{openai_gpt-4o_2024}
OpenAI, A.~Hurst, A.~Lerer, A.~P. Goucher, A.~Perelman, A.~Ramesh, A.~Clark, A.~J. Ostrow, A.~Welihinda \emph{et~al.}, ``{GPT}-4o {System} {Card},'' Oct. 2024, arXiv:2410.21276.

\bibitem[Blattmann et~al.(2023)Blattmann, Dockhorn, Kulal, Mendelevitch, Kilian, Lorenz, Levi, English, Voleti, Letts, Jampani, and Rombach]{blattmann_stable_2023}
A.~Blattmann, T.~Dockhorn, S.~Kulal, D.~Mendelevitch, M.~Kilian, D.~Lorenz, Y.~Levi, Z.~English, V.~Voleti \emph{et~al.}, ``Stable {Video} {Diffusion}: {Scaling} {Latent} {Video} {Diffusion} {Models} to {Large} {Datasets},'' Nov. 2023, arXiv:2311.15127.

\bibitem[Yang et~al.(2024)Yang, Teng, Zheng, Ding, Huang, Xu, Yang, Hong, Zhang, Feng, Yin, Yuxuan.Zhang, Wang, Cheng, Xu, Gu, Dong, and Tang]{yang_cogvideox_2024}
Z.~Yang, J.~Teng, W.~Zheng, M.~Ding, S.~Huang, J.~Xu, Y.~Yang, W.~Hong, X.~Zhang \emph{et~al.}, ``\BIBforeignlanguage{en}{{CogVideoX}: {Text}-to-{Video} {Diffusion} {Models} with {An} {Expert} {Transformer}},'' in \emph{\BIBforeignlanguage{en}{International {Conference} on {Learning} {Representations} ({ICLR})}}, Oct. 2024.

\bibitem[Kong et~al.(2025)Kong, Tian, Zhang, Min, Dai, Zhou, Xiong, Li, Wu, Zhang, Wu, Lin, Yuan, Long, Wang, Wang, Li, Huang, Yang, Tan, Wang, Song, Bai, Wu, Xue, Wang, Wang, Liu, Li, Li, Wang, Yu, Deng, Li, Chen, Cui, Peng, Yu, He, Xu, Zhou, Xu, Tao, Lu, Liu, Zhou, Wang, Yang, Wang, Liu, Jiang, and Zhong]{kong_hunyuanvideo_2025}
W.~Kong, Q.~Tian, Z.~Zhang, R.~Min, Z.~Dai, J.~Zhou, J.~Xiong, X.~Li, B.~Wu \emph{et~al.}, ``{HunyuanVideo}: {A} {Systematic} {Framework} {For} {Large} {Video} {Generative} {Models},'' Mar. 2025, arXiv:2412.03603 [cs].

\bibitem[Wan et~al.(2025)Wan, Wang, Ai, Wen, Mao, Xie, Chen, Yu, Zhao, Yang, Zeng, Wang, Zhang, Zhou, Wang, Chen, Zhu, Zhao, Yan, Huang, Feng, Zhang, Li, Wu, Chu, Feng, Zhang, Sun, Fang, Wang, Gui, Weng, Shen, Lin, Wang, Wang, Zhou, Wang, Shen, Yu, Shi, Huang, Xu, Kou, Lv, Li, Liu, Wang, Zhang, Huang, Li, Wu, Liu, Pan, Zheng, Hong, Shi, Feng, Jiang, Han, Wu, and Liu]{wan_wan_2025}
T.~Wan, A.~Wang, B.~Ai, B.~Wen, C.~Mao, C.-W. Xie, D.~Chen, F.~Yu, H.~Zhao \emph{et~al.}, ``Wan: {Open} and {Advanced} {Large}-{Scale} {Video} {Generative} {Models},'' Apr. 2025, arXiv:2503.20314 [cs].

\bibitem[Stone et~al.(2023)Stone, Xiao, Lu, Gopalakrishnan, Lee, Vuong, Wohlhart, Kirmani, Zitkovich, Xia, Finn, and Hausman]{stone_open-world_2023}
A.~Stone, T.~Xiao, Y.~Lu, K.~Gopalakrishnan, K.-H. Lee, Q.~Vuong, P.~Wohlhart, S.~Kirmani, B.~Zitkovich \emph{et~al.}, ``\BIBforeignlanguage{en}{Open-{World} {Object} {Manipulation} using {Pre}-{Trained} {Vision}-{Language} {Models}},'' in \emph{\BIBforeignlanguage{en}{Conference on {Robot} {Learning} ({CoRL})}}, Dec. 2023.

\bibitem[Dalal et~al.(2025)Dalal, Liu, Talbott, Chen, Pathak, Zhang, and Salakhutdinov]{dalal_local_2025}
M.~Dalal, M.~Liu, W.~Talbott, C.~Chen, D.~Pathak, J.~Zhang, and R.~Salakhutdinov, ``Local {Policies} {Enable} {Zero}-shot {Long}-horizon {Manipulation},'' Mar. 2025, arXiv:2410.22332 [cs].

\bibitem[Patel et~al.(2025)Patel, Mohan, Mai, Jain, Lazebnik, and Li]{patel_robotic_2025}
S.~Patel, S.~Mohan, H.~Mai, U.~Jain, S.~Lazebnik, and Y.~Li, ``Robotic {Manipulation} by {Imitating} {Generated} {Videos} {Without} {Physical} {Demonstrations},'' Jul. 2025, arXiv:2507.00990 [cs].

\bibitem[Yu et~al.(2023)Yu, Gileadi, Fu, Kirmani, Lee, Arenas, Chiang, Erez, Hasenclever, Humplik, Ichter, Xiao, Xu, Zeng, Zhang, Heess, Sadigh, Tan, Tassa, and Xia]{yu_language_2023}
W.~Yu, N.~Gileadi, C.~Fu, S.~Kirmani, K.-H. Lee, M.~G. Arenas, H.-T.~L. Chiang, T.~Erez, L.~Hasenclever \emph{et~al.}, ``\BIBforeignlanguage{en}{Language to {Rewards} for {Robotic} {Skill} {Synthesis}},'' in \emph{\BIBforeignlanguage{en}{Proceedings of {The} 7th {Conference} on {Robot} {Learning}}}.\hskip 1em plus 0.5em minus 0.4em\relax PMLR, Dec. 2023, pp. 374--404, iSSN: 2640-3498.

\bibitem[Goldberg(2025)]{goldberg_good_2025}
K.~Goldberg, ``Good old-fashioned engineering can close the 100,000-year “data gap” in robotics,'' \emph{Science Robotics}, vol.~10, no. 105, p. eaea7390, Aug. 2025.

\bibitem[Black et~al.(2023)Black, Nakamoto, Atreya, Walke, Finn, Kumar, and Levine]{black_zero-shot_2023}
K.~Black, M.~Nakamoto, P.~Atreya, H.~R. Walke, C.~Finn, A.~Kumar, and S.~Levine, ``\BIBforeignlanguage{en}{Zero-{Shot} {Robotic} {Manipulation} with {Pre}-{Trained} {Image}-{Editing} {Diffusion} {Models}},'' in \emph{\BIBforeignlanguage{en}{International {Conference} on {Learning} {Representations} ({ICLR})}}, Oct. 2023.

\bibitem[Bharadhwaj et~al.(2024{\natexlab{a}})Bharadhwaj, Gupta, Kumar, and Tulsiani]{bharadhwaj_towards_2024}
H.~Bharadhwaj, A.~Gupta, V.~Kumar, and S.~Tulsiani, ``Towards {Generalizable} {Zero}-{Shot} {Manipulation} via {Translating} {Human} {Interaction} {Plans},'' in \emph{{IEEE} {International} {Conference} on {Robotics} and {Automation} ({ICRA})}, May 2024.

\bibitem[Huang et~al.(2025{\natexlab{a}})Huang, Wang, Li, Zhang, and Fei-Fei]{huang_rekep_2025}
W.~Huang, C.~Wang, Y.~Li, R.~Zhang, and L.~Fei-Fei, ``\BIBforeignlanguage{en}{{ReKep}: {Spatio}-{Temporal} {Reasoning} of {Relational} {Keypoint} {Constraints} for {Robotic} {Manipulation}},'' in \emph{\BIBforeignlanguage{en}{Conference on {Robot} {Learning} ({CoRL})}}, Jan. 2025.

\bibitem[Yin et~al.(2025{\natexlab{a}})Yin, Li, Wang, McConachie, Shah, Hashimoto, Zhang, Liu, and Li]{yin_codediffuser_2025}
G.~Yin, Y.~Li, Y.~Wang, D.~McConachie, P.~Shah, K.~Hashimoto, H.~Zhang, K.~Liu, and Y.~Li, ``{CodeDiffuser}: {Attention}-{Enhanced} {Diffusion} {Policy} via {VLM}-{Generated} {Code} for {Instruction} {Ambiguity},'' in \emph{Robotics: {Science} and {Systems} ({RSS}) {XXI}}.\hskip 1em plus 0.5em minus 0.4em\relax arXiv, Jun. 2025.

\bibitem[Liang et~al.(2023)Liang, Huang, Xia, Xu, Hausman, Ichter, Florence, and Zeng]{liang_code_2023}
J.~Liang, W.~Huang, F.~Xia, P.~Xu, K.~Hausman, B.~Ichter, P.~Florence, and A.~Zeng, ``Code as {Policies}: {Language} {Model} {Programs} for {Embodied} {Control},'' in \emph{{IEEE} {International} {Conference} on {Robotics} and {Automation} ({ICRA})}, May 2023.

\bibitem[Xu et~al.(2025)Xu, Xu, Xu, Chi, Wetzstein, Veloso, and Song]{xu_flow_2025}
M.~Xu, Z.~Xu, Y.~Xu, C.~Chi, G.~Wetzstein, M.~Veloso, and S.~Song, ``\BIBforeignlanguage{en}{Flow as the {Cross}-domain {Manipulation} {Interface}},'' in \emph{\BIBforeignlanguage{en}{Conference on {Robot} {Learning} ({CoRL})}}, Jan. 2025.

\bibitem[Huang et~al.(2024)Huang, Schmeckpeper, Wang, Biza, Qian, Liu, Jia, Platt, and Walters]{huang_imagination_2024}
H.~Huang, K.~Schmeckpeper, D.~Wang, O.~Biza, Y.~Qian, H.~Liu, M.~Jia, R.~Platt, and R.~Walters, ``{IMAGINATION} {POLICY}: {Using} {Generative} {Point} {Cloud} {Models} for {Learning} {Manipulation} {Policies},'' in \emph{Conference on {Robot} {Learning} ({CoRL})}, Sep. 2024.

\bibitem[Du et~al.(2023)Du, Yang, Dai, Dai, Nachum, Tenenbaum, Schuurmans, and Abbeel]{du_learning_2023}
Y.~Du, S.~Yang, B.~Dai, H.~Dai, O.~Nachum, J.~Tenenbaum, D.~Schuurmans, and P.~Abbeel, ``\BIBforeignlanguage{en}{Learning {Universal} {Policies} via {Text}-{Guided} {Video} {Generation}},'' in \emph{\BIBforeignlanguage{en}{Advances in {Neural} {Information} {Processing} {Systems} ({NeurIPS})}}, Dec. 2023.

\bibitem[Liang et~al.(2025{\natexlab{a}})Liang, Liu, Ozguroglu, Sudhakar, Dave, Tokmakov, Song, and Vondrick]{liang_dreamitate_2025}
J.~Liang, R.~Liu, E.~Ozguroglu, S.~Sudhakar, A.~Dave, P.~Tokmakov, S.~Song, and C.~Vondrick, ``\BIBforeignlanguage{en}{Dreamitate: {Real}-{World} {Visuomotor} {Policy} {Learning} via {Video} {Generation}},'' in \emph{\BIBforeignlanguage{en}{Conference on {Robot} {Learning} ({CoRL})}}, Jan. 2025.

\bibitem[Li et~al.(2025{\natexlab{a}})Li, Gao, Sadigh, and Song]{li_unified_2025}
S.~Li, Y.~Gao, D.~Sadigh, and S.~Song, ``Unified {Video} {Action} {Model},'' Mar. 2025, arXiv:2503.00200 [cs].

\bibitem[Ko et~al.(2023)Ko, Mao, Du, Sun, and Tenenbaum]{ko_learning_2023}
P.-C. Ko, J.~Mao, Y.~Du, S.-H. Sun, and J.~B. Tenenbaum, ``\BIBforeignlanguage{en}{Learning to {Act} from {Actionless} {Videos} through {Dense} {Correspondences}},'' in \emph{\BIBforeignlanguage{en}{International {Conference} on {Learning} {Representations} ({ICLR})}}, Oct. 2023.

\bibitem[Li et~al.(2025{\natexlab{b}})Li, Tucker, Cole, Wang, Jin, Ye, Kanazawa, Holynski, and Snavely]{li_megasam_2025}
Z.~Li, R.~Tucker, F.~Cole, Q.~Wang, L.~Jin, V.~Ye, A.~Kanazawa, A.~Holynski, and N.~Snavely, ``\BIBforeignlanguage{en}{{MegaSaM}: {Accurate}, {Fast} and {Robust} {Structure} and {Motion} from {Casual} {Dynamic} {Videos}},'' in \emph{\BIBforeignlanguage{en}{{IEEE}/{CVF} {Conference} on {Computer} {Vision} and {Pattern} {Recognition} ({CVPR})}}, 2025.

\bibitem[Zhang et~al.(2025{\natexlab{a}})Zhang, Ke, Harley, and Fragkiadaki]{zhang_tapip3d_2025}
B.~Zhang, L.~Ke, A.~W. Harley, and K.~Fragkiadaki, ``{TAPIP3D}: {Tracking} {Any} {Point} in {Persistent} {3D} {Geometry},'' Apr. 2025, arXiv:2504.14717 [cs].

\bibitem[Liu et~al.(2025)Liu, Zeng, Ren, Li, Zhang, Yang, Jiang, Li, Yang, Su, Zhu, and Zhang]{liu_grounding_2025}
S.~Liu, Z.~Zeng, T.~Ren, F.~Li, H.~Zhang, J.~Yang, Q.~Jiang, C.~Li, J.~Yang \emph{et~al.}, ``\BIBforeignlanguage{en}{Grounding {DINO}: {Marrying} {DINO} with {Grounded} {Pre}-training for {Open}-{Set} {Object} {Detection}},'' in \emph{\BIBforeignlanguage{en}{European {Conference} on {Computer} {Vision} ({ECCV})}}, 2025.

\bibitem[Ravi et~al.(2024)Ravi, Gabeur, Hu, Hu, Ryali, Ma, Khedr, Rädle, Rolland, Gustafson, Mintun, Pan, Alwala, Carion, Wu, Girshick, Dollar, and Feichtenhofer]{ravi_sam_2024}
N.~Ravi, V.~Gabeur, Y.-T. Hu, R.~Hu, C.~Ryali, T.~Ma, H.~Khedr, R.~Rädle, C.~Rolland \emph{et~al.}, ``\BIBforeignlanguage{en}{{SAM} 2: {Segment} {Anything} in {Images} and {Videos}},'' in \emph{\BIBforeignlanguage{en}{International {Conference} on {Learning} {Representations} ({ICLR})}}, Oct. 2024.

\bibitem[Kabsch(1976)]{kabsch_solution_1976}
W.~Kabsch, ``\BIBforeignlanguage{en}{A solution for the best rotation to relate two sets of vectors},'' \emph{\BIBforeignlanguage{en}{Acta Crystallographica Section A: Crystal Physics, Diffraction, Theoretical and General Crystallography}}, vol.~32, no.~5, pp. 922--923, Sep. 1976.

\bibitem[Jiang et~al.(2025)Jiang, Hsu, Zhang, Yu, Wang, and Li]{jiang_phystwin_2025}
H.~Jiang, H.-Y. Hsu, K.~Zhang, H.-N. Yu, S.~Wang, and Y.~Li, ``{PhysTwin}: {Physics}-{Informed} {Reconstruction} and {Simulation} of {Deformable} {Objects} from {Videos},'' Mar. 2025, arXiv:2503.17973 [cs].

\bibitem[Zhang et~al.(2025{\natexlab{b}})Zhang, Li, Hauser, and Li]{zhang_particle-grid_2025}
K.~Zhang, B.~Li, K.~Hauser, and Y.~Li, ``Particle-{Grid} {Neural} {Dynamics} for {Learning} {Deformable} {Object} {Models} from {RGB}-{D} {Videos},'' in \emph{Robotics: {Science} and {Systems} ({RSS})}, Jun. 2025.

\bibitem[Lee et~al.(2025)Lee, Duan, Fang, Deng, Liu, Li, Fang, Zhang, Wang, Lee, Han, Pumacay, Wu, Hendrix, Farley, VanderBilt, Farhadi, Fox, and Krishna]{lee_molmoact_2025}
J.~Lee, J.~Duan, H.~Fang, Y.~Deng, S.~Liu, B.~Li, B.~Fang, J.~Zhang, Y.~R. Wang \emph{et~al.}, ``{MolmoAct}: {Action} {Reasoning} {Models} that can {Reason} in {Space},'' Aug. 2025, arXiv:2508.07917 [cs].

\bibitem[Black et~al.(2024)Black, Brown, Driess, Esmail, Equi, Finn, Fusai, Groom, Hausman, Ichter, Jakubczak, Jones, Ke, Levine, Li-Bell, Mothukuri, Nair, Pertsch, Shi, Tanner, Vuong, Walling, Wang, and Zhilinsky]{black_pi_0_2024}
K.~Black, N.~Brown, D.~Driess, A.~Esmail, M.~Equi, C.~Finn, N.~Fusai, L.~Groom, K.~Hausman \emph{et~al.}, ``\${\textbackslash}pi\_0\$: {A} {Vision}-{Language}-{Action} {Flow} {Model} for {General} {Robot} {Control},'' Nov. 2024, arXiv:2410.24164.

\bibitem[Ajay et~al.(2023)Ajay, Han, Du, Li, Gupta, Jaakkola, Tenenbaum, Kaelbling, Srivastava, and Agrawal]{ajay_compositional_2023}
A.~Ajay, S.~Han, Y.~Du, S.~Li, A.~Gupta, T.~Jaakkola, J.~Tenenbaum, L.~Kaelbling, A.~Srivastava, and P.~Agrawal, ``\BIBforeignlanguage{en}{Compositional {Foundation} {Models} for {Hierarchical} {Planning}},'' \emph{\BIBforeignlanguage{en}{Advances in Neural Information Processing Systems}}, vol.~36, pp. 22\,304--22\,325, Dec. 2023.

\bibitem[Bharadhwaj et~al.(2024{\natexlab{b}})Bharadhwaj, Dwibedi, Gupta, Tulsiani, Doersch, Xiao, Shah, Xia, Sadigh, and Kirmani]{bharadhwaj_gen2act_2024}
H.~Bharadhwaj, D.~Dwibedi, A.~Gupta, S.~Tulsiani, C.~Doersch, T.~Xiao, D.~Shah, F.~Xia, D.~Sadigh, and S.~Kirmani, ``{Gen2Act}: {Human} {Video} {Generation} in {Novel} {Scenarios} enables {Generalizable} {Robot} {Manipulation},'' Sep. 2024, arXiv:2409.16283.

\bibitem[Liang et~al.(2025{\natexlab{b}})Liang, Tokmakov, Liu, Sudhakar, Shah, Ambrus, and Vondrick]{liang_video_2025}
J.~Liang, P.~Tokmakov, R.~Liu, S.~Sudhakar, P.~Shah, R.~Ambrus, and C.~Vondrick, ``Video {Generators} are {Robot} {Policies},'' Aug. 2025, arXiv:2508.00795 [cs].

\bibitem[Yuan et~al.(2024)Yuan, Wen, Zhang, and Gao]{yuan_general_2024}
C.~Yuan, C.~Wen, T.~Zhang, and Y.~Gao, ``General {Flow} as {Foundation} {Affordance} for {Scalable} {Robot} {Learning},'' Jan. 2024, arXiv:2401.11439 [cs].

\bibitem[Eisner et~al.(2024)Eisner, Zhang, and Held]{eisner_flowbot3d_2024}
B.~Eisner, H.~Zhang, and D.~Held, ``{FlowBot3D}: {Learning} {3D} {Articulation} {Flow} to {Manipulate} {Articulated} {Objects},'' May 2024, arXiv:2205.04382 [cs].

\bibitem[Yin et~al.(2025{\natexlab{b}})Yin, Yang, and Abbeel]{yin_object-centric_2025}
Z.-H. Yin, S.~Yang, and P.~Abbeel, ``Object-centric {3D} {Motion} {Field} for {Robot} {Learning} from {Human} {Videos},'' Jun. 2025, arXiv:2506.04227 [cs].

\bibitem[Bharadhwaj et~al.(2024{\natexlab{c}})Bharadhwaj, Mottaghi, Gupta, and Tulsiani]{bharadhwaj_track2act_2024}
H.~Bharadhwaj, R.~Mottaghi, A.~Gupta, and S.~Tulsiani, ``\BIBforeignlanguage{en}{{Track2Act}: {Predicting} {Point} {Tracks} from {Internet} {Videos} {Enables} {Generalizable} {Robot} {Manipulation}},'' in \emph{\BIBforeignlanguage{en}{European {Conference} on {Computer} {Vision} ({ECCV})}}, 2024.

\bibitem[Zhang et~al.(2024{\natexlab{a}})Zhang, Eisner, and Held]{zhang_flowbot_2024}
H.~Zhang, B.~Eisner, and D.~Held, ``{FlowBot}++: {Learning} {Generalized} {Articulated} {Objects} {Manipulation} via {Articulation} {Projection},'' May 2024, arXiv:2306.12893 [cs].

\bibitem[Chen et~al.(2025)Chen, Sun, Zhang, Pollefeys, and Leutenegger]{chen_vidbot_2025}
H.~Chen, B.~Sun, A.~Zhang, M.~Pollefeys, and S.~Leutenegger, ``\BIBforeignlanguage{en}{{VidBot}: {Learning} {Generalizable} {3D} {Actions} from {In}-the-{Wild} {2D} {Human} {Videos} for {Zero}-{Shot} {Robotic} {Manipulation}},'' in \emph{\BIBforeignlanguage{en}{{IEEE}/{CVF} {Conference} on {Computer} {Vision} and {Pattern} {Recognition} ({CVPR})}}, 2025.

\bibitem[Zhi et~al.(2025)Zhi, Chen, Zhou, Dong, Wu, Han, and Tan]{zhi_3dflowaction_2025}
H.~Zhi, P.~Chen, S.~Zhou, Y.~Dong, Q.~Wu, L.~Han, and M.~Tan, ``{3DFlowAction}: {Learning} {Cross}-{Embodiment} {Manipulation} from {3D} {Flow} {World} {Model},'' Jun. 2025, arXiv:2506.06199.

\bibitem[Shi et~al.(2025)Shi, Zhao, Wang, Pedroza, Luo, Wang, Ma, and Jayaraman]{shi_zeromimic_2025}
J.~Shi, Z.~Zhao, T.~Wang, I.~Pedroza, A.~Luo, J.~Wang, J.~Ma, and D.~Jayaraman, ``{ZeroMimic}: {Distilling} {Robotic} {Manipulation} {Skills} from {Web} {Videos},'' Mar. 2025, arXiv:2503.23877 [cs].

\bibitem[Zhou et~al.(2025)Zhou, Pan, LeCun, and Pinto]{zhou_dino-wm_2025}
G.~Zhou, H.~Pan, Y.~LeCun, and L.~Pinto, ``\BIBforeignlanguage{en}{{DINO}-{WM}: {World} {Models} on {Pre}-trained {Visual} {Features} enable {Zero}-shot {Planning}},'' in \emph{\BIBforeignlanguage{en}{International {Conference} on {Machine} {Learning} ({ICML})}}, Jun. 2025.

\bibitem[Labs et~al.(2025)Labs, Batifol, Blattmann, Boesel, Consul, Diagne, Dockhorn, English, English, Esser, Kulal, Lacey, Levi, Li, Lorenz, Müller, Podell, Rombach, Saini, Sauer, and Smith]{labs_flux1_2025}
B.~F. Labs, S.~Batifol, A.~Blattmann, F.~Boesel, S.~Consul, C.~Diagne, T.~Dockhorn, J.~English, Z.~English \emph{et~al.}, ``{FLUX}.1 {Kontext}: {Flow} {Matching} for {In}-{Context} {Image} {Generation} and {Editing} in {Latent} {Space},'' Jun. 2025, arXiv:2506.15742 [cs].

\bibitem[Murali et~al.(2025)Murali, Sundaralingam, Chao, Yuan, Yamada, Carlson, Ramos, Birchfield, Fox, and Eppner]{murali_graspgen_2025}
A.~Murali, B.~Sundaralingam, Y.-W. Chao, W.~Yuan, J.~Yamada, M.~Carlson, F.~Ramos, S.~Birchfield, D.~Fox, and C.~Eppner, ``{GraspGen}: {A} {Diffusion}-based {Framework} for 6-{DOF} {Grasping} with {On}-{Generator} {Training},'' Jul. 2025, arXiv:2507.13097.

\bibitem[Zhang et~al.(2024{\natexlab{b}})Zhang, Li, Hauser, and Li]{zhang_adaptigraph_2024}
K.~Zhang, B.~Li, K.~Hauser, and Y.~Li, ``{AdaptiGraph}: {Material}-{Adaptive} {Graph}-{Based} {Neural} {Dynamics} for {Robotic} {Manipulation},'' in \emph{Robotics: {Science} and {Systems} ({RSS})}, vol.~20, Jul. 2024.

\bibitem[Wang et~al.(2023)Wang, Li, Driggs-Campbell, Fei-Fei, and Wu]{wang_dynamic-resolution_2023}
Y.~Wang, Y.~Li, K.~Driggs-Campbell, L.~Fei-Fei, and J.~Wu, ``Dynamic-{Resolution} {Model} {Learning} for {Object} {Pile} {Manipulation},'' in \emph{Robotics: {Science} and {Systems} ({RSS})}, Jul. 2023.

\bibitem[Huang et~al.(2025{\natexlab{b}})Huang, Chen, Zhang, Sun, and Schwager]{huang_particleformer_2025}
S.~Huang, Q.~Chen, X.~Zhang, J.~Sun, and M.~Schwager, ``{ParticleFormer}: {A} {3D} {Point} {Cloud} {World} {Model} for {Multi}-{Object}, {Multi}-{Material} {Robotic} {Manipulation},'' in \emph{Conference on {Robot} {Learning} ({CoRL})}, Jul. 2025.

\bibitem[Chi et~al.(2023)Chi, Feng, Du, Xu, Cousineau, Burchfiel, and Song]{chi_diffusion_2023}
C.~Chi, S.~Feng, Y.~Du, Z.~Xu, E.~Cousineau, B.~C. Burchfiel, and S.~Song, ``Diffusion {Policy}: {Visuomotor} {Policy} {Learning} via {Action} {Diffusion},'' vol.~19, Jul. 2023.

\bibitem[Ren et~al.(2024)Ren, Liu, Zeng, Lin, Li, Cao, Chen, Huang, Chen, Yan, Zeng, Zhang, Li, Yang, Li, Jiang, and Zhang]{ren_grounded_2024}
T.~Ren, S.~Liu, A.~Zeng, J.~Lin, K.~Li, H.~Cao, J.~Chen, X.~Huang, Y.~Chen \emph{et~al.}, ``Grounded {SAM}: {Assembling} {Open}-{World} {Models} for {Diverse} {Visual} {Tasks},'' Jan. 2024, arXiv:2401.14159 [cs].

\bibitem[Wiedemer et~al.(2025)Wiedemer, Li, Vicol, Gu, Matarese, Swersky, Kim, Jaini, and Geirhos]{wiedemer_video_2025}
T.~Wiedemer, Y.~Li, P.~Vicol, S.~S. Gu, N.~Matarese, K.~Swersky, B.~Kim, P.~Jaini, and R.~Geirhos, ``Video models are zero-shot learners and reasoners,'' Sep. 2025, arXiv:2509.20328 [cs].

\bibitem[Wang et~al.(2024)Wang, Xu, Dai, Xiang, Deng, Tong, and Yang]{wang_moge_2024}
R.~Wang, S.~Xu, C.~Dai, J.~Xiang, Y.~Deng, X.~Tong, and J.~Yang, ``{MoGe}: {Unlocking} {Accurate} {Monocular} {Geometry} {Estimation} for {Open}-{Domain} {Images} with {Optimal} {Training} {Supervision},'' Nov. 2024, arXiv:2410.19115 [cs].

\bibitem[Luo et~al.(2020)Luo, Huang, Szeliski, Matzen, and Kopf]{luo_consistent_2020}
X.~Luo, J.-B. Huang, R.~Szeliski, K.~Matzen, and J.~Kopf, ``Consistent video depth estimation,'' \emph{ACM Trans. Graph.}, vol.~39, no.~4, pp. 71:71:1--71:71:13, Aug. 2020.

\bibitem[Zhang et~al.(2022)Zhang, Cole, Li, Rubinstein, Snavely, and Freeman]{zhang_structure_2022}
Z.~Zhang, F.~Cole, Z.~Li, M.~Rubinstein, N.~Snavely, and W.~T. Freeman, ``\BIBforeignlanguage{en}{Structure and {Motion} from {Casual} {Videos}},'' in \emph{\BIBforeignlanguage{en}{Computer {Vision} – {ECCV} 2022}}, S.~Avidan, G.~Brostow, M.~Cissé, G.~M. Farinella, and T.~Hassner, Eds.\hskip 1em plus 0.5em minus 0.4em\relax Cham: Springer Nature Switzerland, 2022, pp. 20--37.

\bibitem[Kim et~al.(2025{\natexlab{b}})Kim, Yi, Choi, Ma, Goldberg, and Kanazawa]{kim_pyroki_2025}
C.~M. Kim, B.~Yi, H.~Choi, Y.~Ma, K.~Goldberg, and A.~Kanazawa, ``{PyRoki}: {A} {Modular} {Toolkit} for {Robot} {Kinematic} {Optimization},'' May 2025, arXiv:2505.03728 [cs].

\end{thebibliography}
\begin{appendices}

\section{Video Generation}
\label{sec:appendix_video_generation}

In this section, we provide additional details on the video generation models, prompt engineering techniques, and the specific prompts used in our experiments.

\subsection{Wan2.1}
Wan2.1 is the latest open-source video generation model from Alibaba at the time of this work.
We note that a newer version, Wan2.2, was released recently but does not support all the modes of Wan2.1 that we require.
Following the official recommendation, we use Chinese prompts, which we found to yield better results than English prompts.

For Wan2.1, we use its Image-to-Video (I2V) model for standard video generation and its First-Last-Frame-to-Video (FLF2V) model when conditioning on a goal image.
We generate 41 frames for each video at a resolution of 1280$\times$720 and a frame rate of 16 FPS. We use their UniPC sampler with 40 sampling steps, a noise shift parameter of 5.0, and a guidance scale of 5.0.

\subsection{Veo}
We also experimented with Veo, a closed-source model from Google. At the time of our experiments, the model supported I2V generation but not goal-image conditioning.
Specifically, we used the \texttt{veo-3.0-generate-001} model via the Vertex AI API.
We generated 8-second videos\footnote{During our experiments, the model only supported 8-second video generation. It now also supports durations of 4 and 6 seconds.} at a resolution of 1280$\times$720 and a frame rate of 24 FPS.
To maintain consistency with Wan2.1, we downsampled the generated videos to 41 frames.

Pricing for the Veo model is subject to change.
At the time of writing, the cost was \$0.20 per second of generated video (e.g., \$1.60 for an 8-second clip).

\subsection{Prompt Engineering}
To improve the quality and controllability of the generated videos, we employ prompt extension, a technique where a simple instruction is automatically enriched with additional details about style, composition, and action.

For Wan2.1, we adapt its official prompt extension script. We use the prompt template from the \href{https://github.com/Wan-Video/Wan2.1/blob/main/wan/utils/prompt_extend.py}{official repository} and pass it to the Gemini 2.5 Pro model to generate the extended Chinese prompt.

For Veo, we utilize its native prompt enhancement feature available through the Vertex AI API, which automatically refines the input prompt for improved generation quality.

\subsection{Generation Prompts}
Below, we provide the original and extended prompts used for each task, along with the corresponding initial image from the robot's perspective.

\begin{figure}[h!]
\centering
\includegraphics[width=\linewidth]{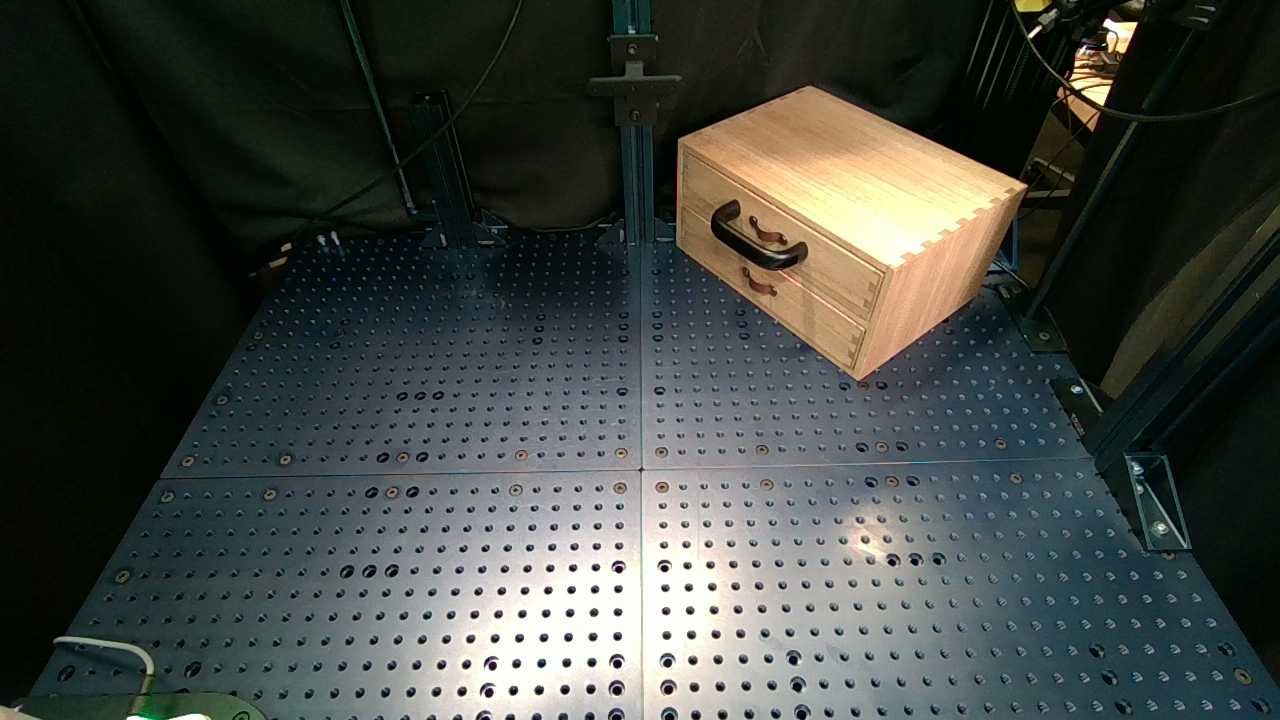}
\caption{Initial observation of the drawer open task.}
\label{fig:initial_observation_drawer_open}
\end{figure}

\begin{tcolorbox}[colback=gray!10,colframe=purple,title=Open Drawer (Wan2.1 original),width=\columnwidth]
\begin{CJK*}{UTF8}{gbsn}
一只人手抓住黑色抽屉把手，顺利地将其从抽屉中拉出。抽屉应沿直线打开，且不会前后移动。人手不会在视觉上遮挡抽屉把手。    
\end{CJK*}
\end{tcolorbox}

\begin{tcolorbox}[colback=gray!10,colframe=purple,title=Open Drawer (Wan2.1 extended),width=\columnwidth]
\begin{CJK*}{UTF8}{gbsn}
实景拍摄，一只人手抓住黑色抽屉把手，将木质抽屉从木制抽屉箱中顺利拉出。抽屉沿着滑轨直线打开，抽屉把手光滑，人手并未遮挡。整体画面为抽屉箱被固定在带有圆形孔洞的蓝色实验平台上，背景是黑色的幕布和金属支架。展示抽屉打开的全过程。   
\end{CJK*}
\end{tcolorbox}

\begin{tcolorbox}[colback=gray!10,colframe=purple,title=Open Drawer (Veo original),width=\columnwidth]
A human hand grasps a black drawer handle and smoothly pulls the drawer out. The drawer should open in a straight line without moving forward or backward. The human hand should not visually obscure the drawer handle.
\end{tcolorbox}

\newpage

\begin{figure}[h!]
\centering
\includegraphics[width=\linewidth]{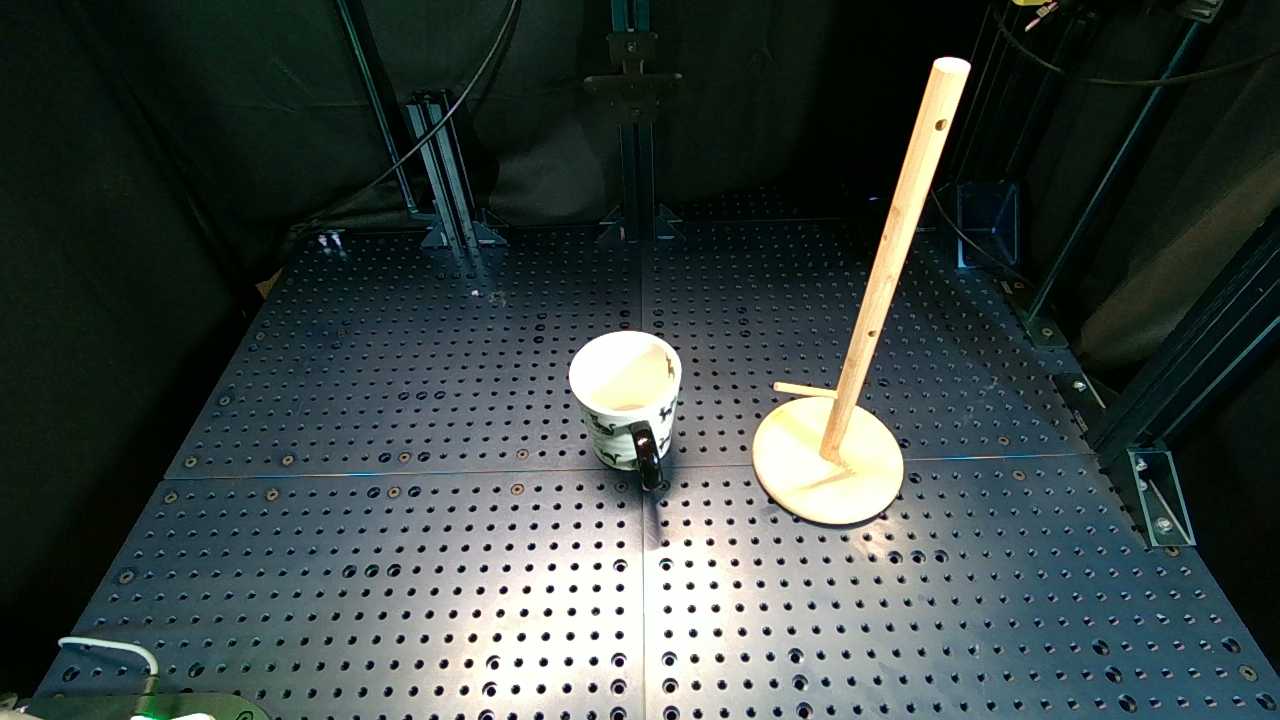}
\caption{Initial observation of the hang mug task.}
\label{fig:initial_observation_hang_mug}
\end{figure}

\begin{tcolorbox}[colback=gray!10,colframe=purple,title=Hang Mug (Wan2.1 original),width=\columnwidth]
\begin{CJK*}{UTF8}{gbsn}
一只人的手拿起杯子并将其挂在木架上。人的手不会在视觉上遮挡杯子。
\end{CJK*}
\end{tcolorbox}

\begin{tcolorbox}[colback=gray!10,colframe=purple,title=Hang Mug (Wan2.1 extended),width=\columnwidth]
\begin{CJK*}{UTF8}{gbsn}
专业工作室摄影，一只手正将一个印有熊猫图案的白色纸杯挂到右侧的木质支架上。纸杯内的液体在过程中有轻微晃动。镜头保持固定，只对物体进行平移，从杯子被拿起，到杯子稳稳地挂在支架上。画面呈现高对比度的冷色调，强调了物品的质感和细节。整体为中景拍摄，突出物体的主体性。
\end{CJK*}
\end{tcolorbox}

\begin{tcolorbox}[colback=gray!10,colframe=purple,title=Hang Mug (Veo original),width=\columnwidth]
A human hand picks up the cup and hangs it on the wooden stand. The human hand does not visually obstruct the cup.
\end{tcolorbox}

\newpage

\begin{figure}[h!]
\centering
\includegraphics[width=\linewidth]{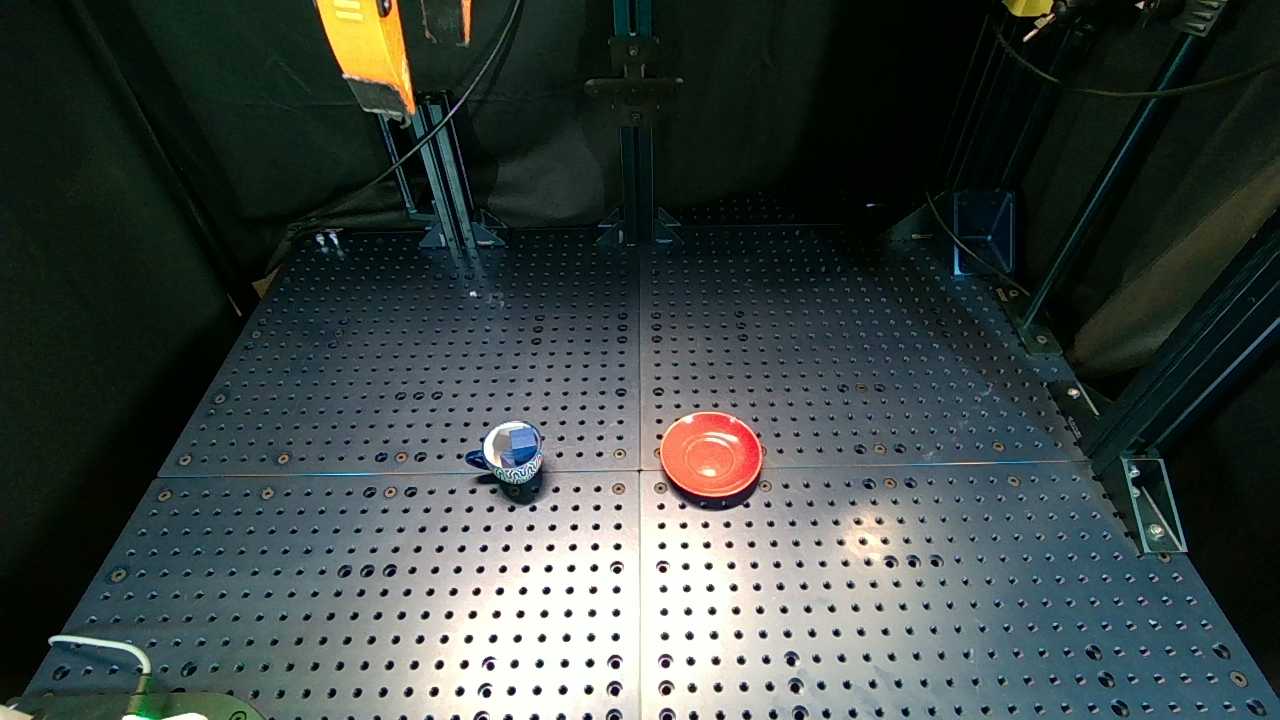}
\caption{Initial observation of the cup on saucer task.}
\label{fig:initial_observation_cup_on_saucer}
\end{figure}

\begin{tcolorbox}[colback=gray!10,colframe=purple,title=Cup on Saucer (Wan2.1 original),width=\columnwidth]
\begin{CJK*}{UTF8}{gbsn}
一只人手拿起蓝色的小杯子，举起来，轻轻地放在它红色盘子上。
\end{CJK*}
\end{tcolorbox}

\begin{tcolorbox}[colback=gray!10,colframe=purple,title=Cup on Saucer (Wan2.1 extended),width=\columnwidth]
\begin{CJK*}{UTF8}{gbsn}
写实风格，一只人手伸出，拿起蓝色带有白色图案的陶瓷小杯子，杯子被稳稳握住。随后，人手将杯子向上举起，并缓慢移动至右侧，将杯子稳稳地放置在一个红色的圆形盘子上。整个过程平稳流畅，镜头从俯视角度拍摄，展现了物体的精细细节和整体的摆放过程。
\end{CJK*}
\end{tcolorbox}

\begin{tcolorbox}[colback=gray!10,colframe=purple,title=Cup on Saucer (Veo original),width=\columnwidth]
A human hand picks up the small blue cup, lifts it up, and gently places it on its red plate.
\end{tcolorbox}

\newpage

\begin{figure}[h!]
\centering
\includegraphics[width=\linewidth]{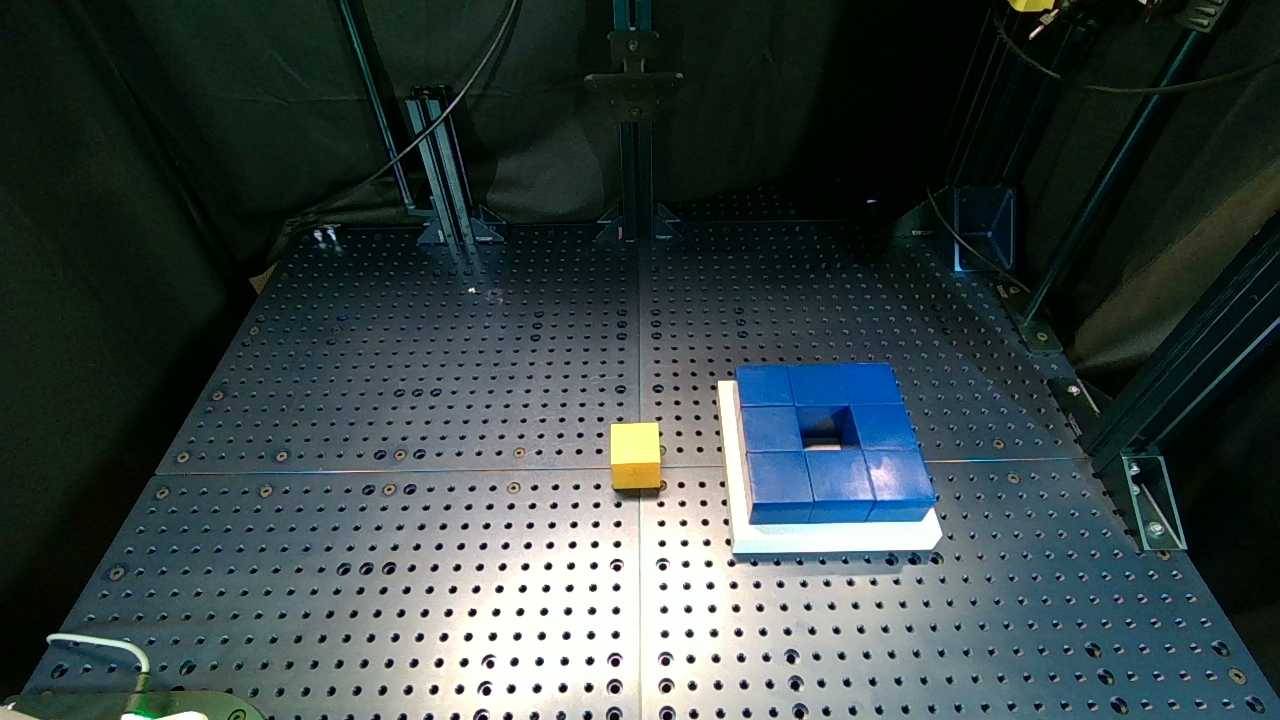}
\caption{Initial observation of the block insertion task.}
\label{fig:initial_observation_block_insertion}
\end{figure}

\begin{tcolorbox}[colback=gray!10,colframe=purple,title=Block Insertion (Wan2.1 original),width=\columnwidth]
\begin{CJK*}{UTF8}{gbsn}
一只人手拿起黄色块，并将其准确地插入有蓝色块的盘子的中心。黄色块应该先上升然后下降。人手不会在视觉上遮挡黄色块。
\end{CJK*}
\end{tcolorbox}

\begin{tcolorbox}[colback=gray!10,colframe=purple,title=Block Insertion (Wan2.1 extended),width=\columnwidth]
\begin{CJK*}{UTF8}{gbsn}
超写实主义近景镜头，一只人手拿起一个黄色方块，然后将它精准地放置在右侧一个由蓝色方块组成的盘子中心。该动作从黄色方块先向上移动，再向下插入盘子中心开始。整个过程中，人手始终保持在黄色方块上方，避免遮挡。画面背景是深色绒布，下方是带有规则孔洞的金属实验平台。镜头稳定，画面清晰。
\end{CJK*}
\end{tcolorbox}

\begin{tcolorbox}[colback=gray!10,colframe=purple,title=Block Insertion (Veo original),width=\columnwidth]
A human hand picks up the yellow block and inserts it precisely into the center of the plate with the blue block. The yellow block should rise first and then fall. The human hand should not visually obscure the yellow block.
\end{tcolorbox}

\newpage

\begin{figure}[h!]
\centering
\includegraphics[width=\linewidth]{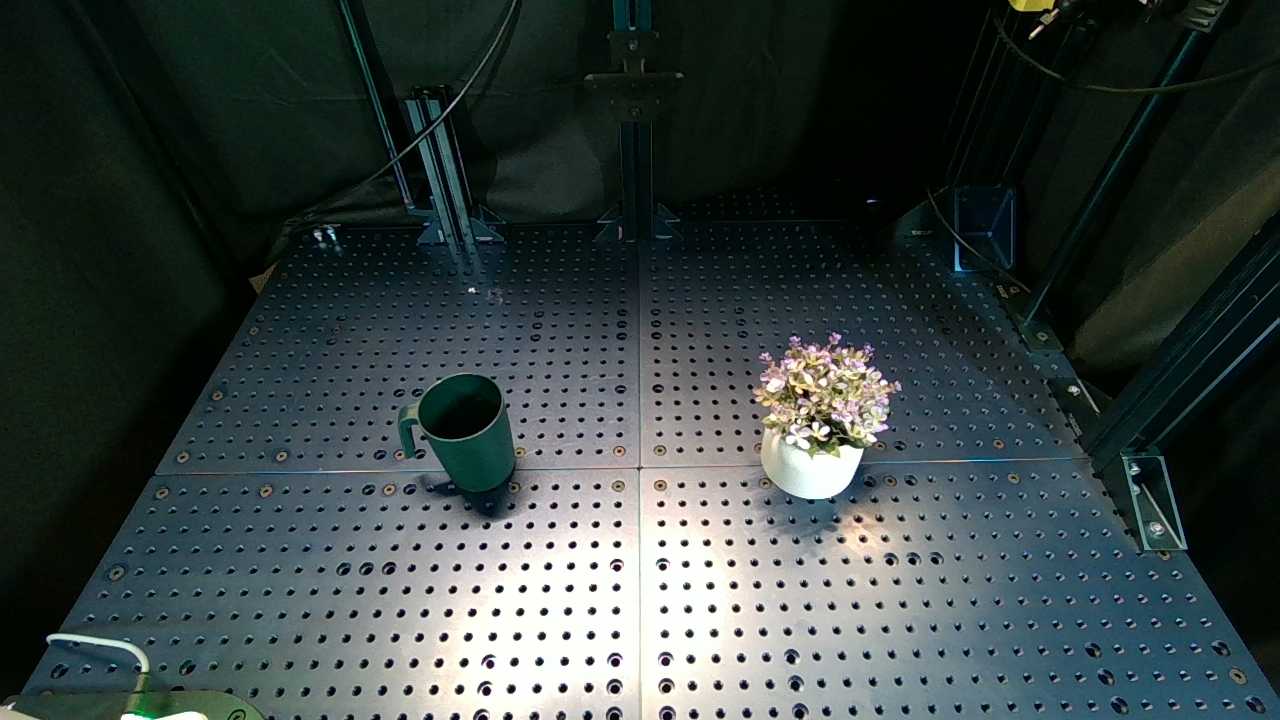}
\caption{Initial observation of the water plant task.}
\label{fig:initial_observation_water_plant}
\end{figure}

\begin{tcolorbox}[colback=gray!10,colframe=purple,title=Water Plant (Wan2.1 original),width=\columnwidth]
\begin{CJK*}{UTF8}{gbsn}
一只人手抓住左边的绿色水杯，将其举起，然后平稳地给植物浇水。摄像机始终保持静止。人手不会在视觉上遮挡杯子。
\end{CJK*}
\end{tcolorbox}

\begin{tcolorbox}[colback=gray!10,colframe=purple,title=Water Plant (Wan2.1 extended),width=\columnwidth]
\begin{CJK*}{UTF8}{gbsn}
仰视视角，一只手持绿色马克杯，马克杯倾斜，准备向白盆里的紫色小花浇水。背景为深邃的黑色布景，地面为带有网格的金属平台，光线集中在平台中央，形成明暗对比。整体画面呈现出一种科技感和实验性。
\end{CJK*}
\end{tcolorbox}

\begin{tcolorbox}[colback=gray!10,colframe=purple,title=Water Plant (Veo original),width=\columnwidth]
A human hand grasps the green water cup on the left, lifts it, and steadily waters the plant. The camera remains stationary throughout. The hand does not visually obscure the cup.
\end{tcolorbox}

\newpage

\begin{figure}[h!]
\centering
\includegraphics[width=\linewidth]{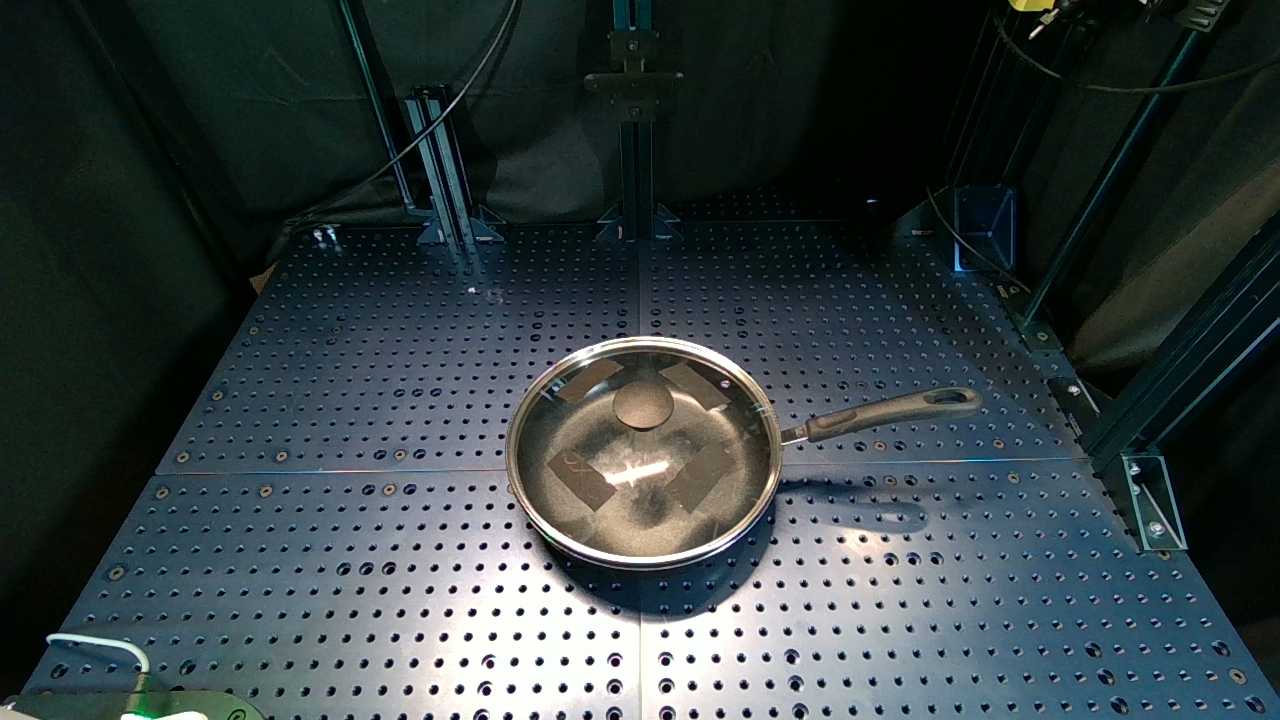}
\caption{Initial observation of the open lid task.}
\label{fig:initial_observation_open_lid}
\end{figure}

\begin{tcolorbox}[colback=gray!10,colframe=purple,title=Open Lid (Wan2.1 original),width=\columnwidth]
\begin{CJK*}{UTF8}{gbsn}
一只人手抓住锅的透明盖子并将其直接提起。摄像机始终保持静止。人手不会在视觉上遮挡盖子。
\end{CJK*}
\end{tcolorbox}

\begin{tcolorbox}[colback=gray!10,colframe=purple,title=Open Lid (Wan2.1 extended),width=\columnwidth]
\begin{CJK*}{UTF8}{gbsn}
写实主义摄影，一只手在抓住平底锅的透明锅盖，并将其提起。锅盖上贴有三张长方形的黑色不干胶，上面有白色的反光。锅盖中间有一个黑色的圆形把手。锅的四周是深蓝色的实验台，表面布满了规则排列的圆孔。背景是深色的幕布和金属支架，光线集中在锅具上，营造出一种局部照明的氛围。近景，固定镜头，俯视角度。
\end{CJK*}
\end{tcolorbox}

\begin{tcolorbox}[colback=gray!10,colframe=purple,title=Open Lid (Veo original),width=\columnwidth]
A human hand grasps the transparent lid of a pot and lifts it straight up. The camera remains stationary. The hand does not visually obscure the lid.
\end{tcolorbox}

\newpage

\begin{figure}[h!]
\centering
\includegraphics[width=\linewidth]{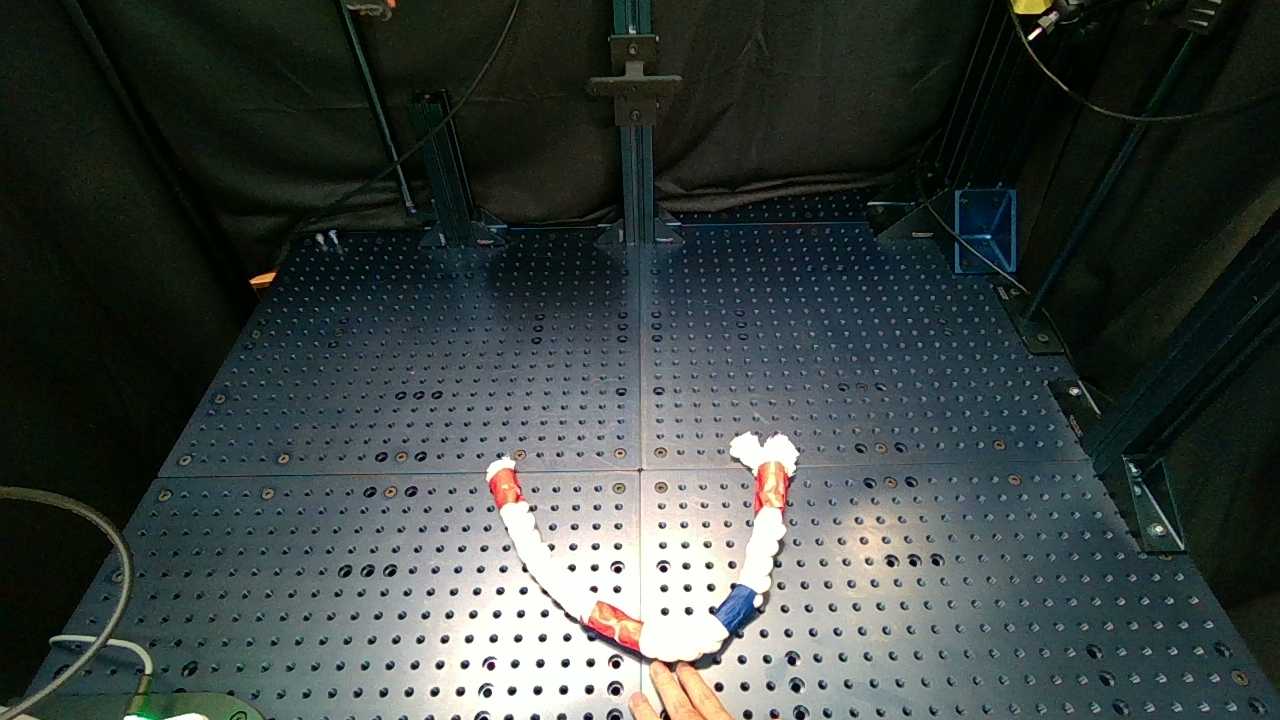}
\caption{Initial observation of the straighten rope task.}
\label{fig:initial_observation_straighten_rope}
\end{figure}

\begin{tcolorbox}[colback=gray!10,colframe=purple,title=Straighten Rope (Wan2.1 original),width=\columnwidth]
\begin{CJK*}{UTF8}{gbsn}
一只人手缓缓地把弯曲的绳子推成直的。
\end{CJK*}
\end{tcolorbox}

\begin{tcolorbox}[colback=gray!10,colframe=purple,title=Straighten Rope (Wan2.1 extended),width=\columnwidth]
\begin{CJK*}{UTF8}{gbsn}
工业风写实记录，在布满孔洞的深色金属实验台上，一只戴着黑色智能手表的人手缓缓出现，将一根红白蓝三色相间的粗布绳从U形弯曲状态，慢慢地推动、抚平成一条直线。整个动作流畅而稳定。镜头固定，采用俯视视角，记录了这一精准的操作过程。
\end{CJK*}
\end{tcolorbox}

\begin{tcolorbox}[colback=gray!10,colframe=purple,title=Straighten Rope (Veo original),width=\columnwidth]
A hand slowly pushes the bent rope into a straight line.
\end{tcolorbox}

\begin{figure*}[t!]
    \centering
    \includegraphics[width=\linewidth]{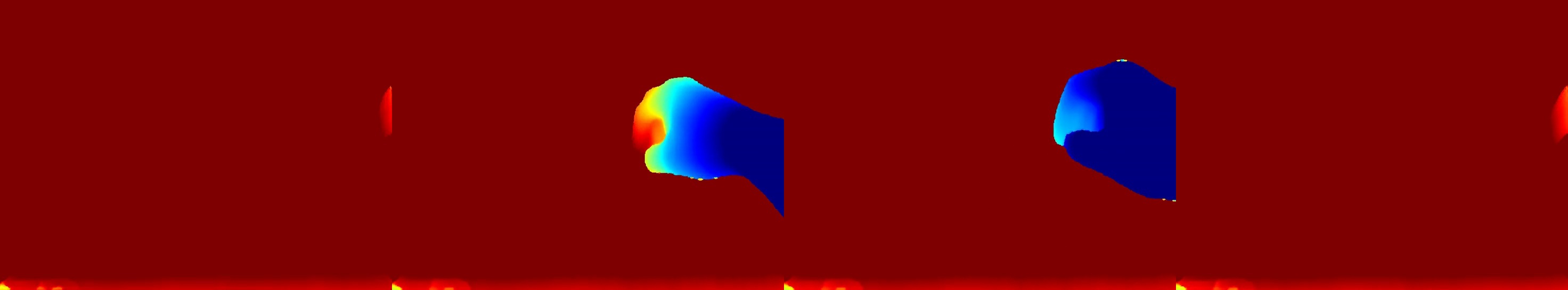}
    \includegraphics[width=\linewidth]{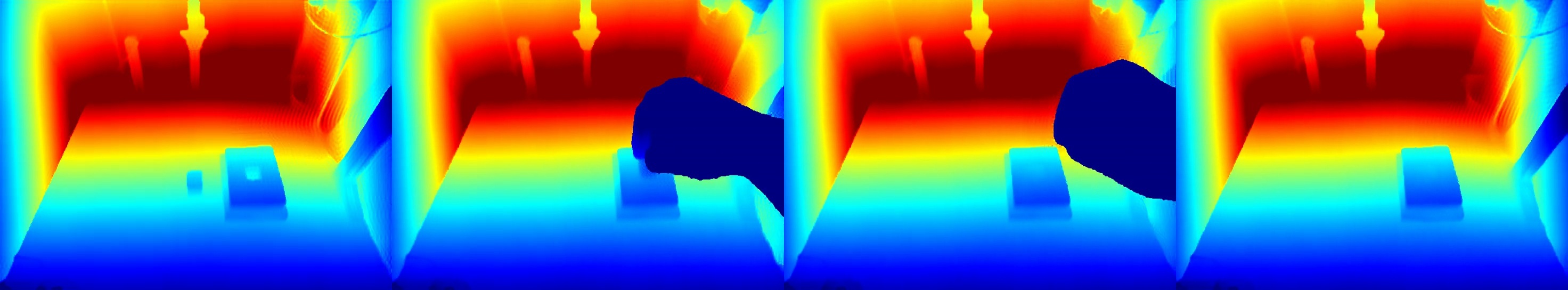}
    \caption{
        \textbf{Comparison of the depth maps before (top) and after (bottom) scaling}.
        The depth maps are visualized in the \texttt{jet} color map, where the colormap range is obtained from the ground-truth depth map.
    }
    \label{fig:depth_maps_before_after_scaling}
\end{figure*}
\newpage

\section{Depth Estimation}
We use the implementation of MegaSaM~\cite{li_megasam_2025} from TAPIP3D~\cite{zhang_tapip3d_2025} for depth estimation.
Specifically, we use the MoGe~\cite{wang_moge_2024} model to estimate the per-frame depth map.
Instead of estimating the camera intrinsics, we use the ground-truth calibrated intrinsics from our camera.
The estimated depth maps are then postprocessed by bundle adjustment and consistent video depth (CVD)~\cite{luo_consistent_2020} optimization following CausalSAM~\cite{zhang_structure_2022}.

Even after these postprocessing steps, the estimated metric depth maps are still ambiguous.
Therefore, we opt to use the initial ground-truth depth map as the reference depth map for calibration.
Specifically, we compute the scaling factor between the median depth of the first estimated frame and the initial ground-truth depth map.
We then multiply the estimated depth maps by this scaling factor to obtain the calibrated depth maps.

After applying the scaling factor (see \cref{fig:depth_maps_before_after_scaling}), the calibrated depth maps are more consistent with the ground-truth depth map and more temporally consistent.
We find the calibrated depth maps are accurate enough to support precise manipulation tasks such as the block insertion as demonstrated in the figure, which requires millimeter-level precision.

\section{3D Point Tracking}
We leverage TAPIP3D~\cite{zhang_tapip3d_2025} for 3D point tracking, which tracks 3D points in the XYZ 3D coordinate space instead of the UVD 2D space.
We generate query points on the first frame using uniform grid sampling of $32 \times 32$ points.
We set the tracker iterations to 6.

\section{Object Grounding}
The previous step produces dense 3D tracking for the entire image.
We need to ground the points to the target object.
We use the Grounded-SAM2 pipeline~\cite{ren_grounded_2024} for object grounding, which combines Grounding DINO~\cite{liu_grounding_2025} and SAM2~\cite{ravi_sam_2024}.
We pass the query object name to the pipeline to extract the object mask throughout the video.
Then, we use the mask video to filter the 3D tracking points and only keep the points that are visible throughout the video.
We set the bounding box threshold to 0.25 and text threshold to 0.3.
We select the bounding box from Grounding DINO with the highest score and set it as the input prompt to SAM2 to extract the object mask.

\section{Rejection Sampling}
After obtaining the 3D object flow, we can project the 3D flow onto the first frame to obtain the 2D object flow.
We then pass object flow images to Gemini 2.5 Pro to filter out hallucinations, such as generative artifacts and implausible motions, that may be unavoidably introduced by the video generation model.
We find this strategy to be effective and benefits from scaling the execution-time computation resources.
For example, we can generate 8 candidates in parallel and select the best one from the 8 candidates using 8 H100 GPUs.

Along with the following system prompt, we also provide the flow image and the task description used to generate the video to clarify our expectation of the task.
\begin{tcolorbox}[colback=gray!10,colframe=purple,title=Rejection Sampling System Prompt,width=\columnwidth]
    You are a flow analysis expert. Analyze the stitched flow image and evaluate which flow visualization (marked with IDs in the top-left corner) represents the most reasonable and natural object motion. 
    The flow is a manual annotation overlay on the image to indicate the intended object motion.
    Consider:
    1. Continuity and smoothness of the flow
    2. Natural motion patterns
    3. Proper object identification (avoid flows that spread throughout the entire image)
    4. Alignment with the task requirements
    You should reject images that show flows throughout the image (which means the object is not identified).
    Provide a clear recommendation on which flow ID is best and why.
\end{tcolorbox}

\begin{figure*}[t!]
    \centering
    \includegraphics[width=\linewidth]{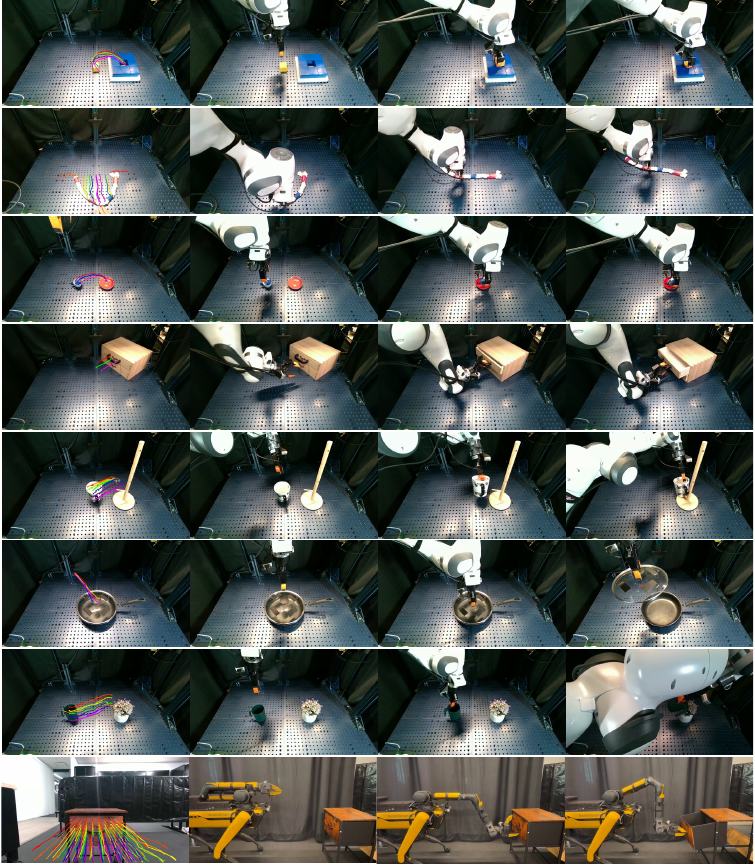}
    \caption{
        \textbf{Real-world manipulation experiments.}
        From top to bottom: block insertion, rope straightening, cup on saucer, open drawer, hang mug, open lid, water plant, and open drawer using the Spot.
    }
    \label{fig:real_world_experiments_appendix}
\end{figure*}

\section{Trajectory Optimization}

During execution time, we refine the sequence of actions using trajectory optimization to find an optimal, collision-free, and smooth sequence of joint configurations $Q = \{q_0, q_1, \dots, q_{T-1}\}$. The trajectory is initialized by linearly interpolating between start and end configurations, $q_{\text{start,IK}}$ and $q_{\text{end,IK}}$, which are pre-calculated using an IK solver. The optimal trajectory $Q^*$ is found by solving the following constrained non-linear optimization problem:
\begin{equation}
\label{eq:appendix_ik_optimization}
\begin{aligned}
\min_{Q} \quad & w_s \mathcal{C}_{\text{smooth}} + w_r \mathcal{C}_{\text{rest}}, \text{\quad subject to} \\
\quad & q_0 = q_{\text{start,IK}} \quad \text{and} \quad q_{T-1} = q_{\text{end,IK}}, \\
& q_{\min} \le q_t \le q_{\max}, \quad \forall t \in \{0, \dots, T-1\}, \\
& d_s(q_t, q_{t+1}, O_j) \ge \epsilon_{\text{safe}}, \quad \forall t, \forall O_j \in \text{Obstacles}.
\end{aligned}
\end{equation}
This optimization problem is solved using a non-linear least-squares algorithm (Levenberg-Marquardt). The constraints for joint limits and collision avoidance are incorporated as high-weight penalty terms in the objective function, while start and end configurations are hard constraints.

\subsection*{Objective and Penalty Terms}

\textbf{Smoothness Cost ($\mathcal{C}_{\text{smooth}}$)}. This cost penalizes non-smooth motion by minimizing the squared norms of joint velocity ($\dot{q}$), which is approximated using finite differences. In practice, this is implemented as a cost on the deviation from the previous joint configuration, encouraging temporal smoothness:
\begin{equation} \mathcal{C}_{\text{smooth}} = \sum_{t} w_s \|q_t - q_{t-1}\|^2. \end{equation}

\textbf{Rest Position Cost ($\mathcal{C}_{\text{rest}}$)}. This is a regularization term that encourages the trajectory to remain close to a default home configuration, $q_{\text{rest}}$:
\begin{equation} \mathcal{C}_{\text{rest}} = \sum_{t} w_r \|q_t - q_{\text{rest}}\|^2. \end{equation}

\textbf{Joint Limits Penalty}. This term penalizes any violation of the minimum ($q_{\min}$) and maximum ($q_{\max}$) joint limits:
\begin{equation}
\begin{split}
\mathcal{C}_{\text{limits}} = \sum_{t} w_l \big( &\|\max(0, q_t - q_{\max})\|^2 \\
& + \|\max(0, q_{\min} - q_t)\|^2 \big).
\end{split}
\end{equation}

\textbf{Collision Avoidance Penalty}. This term enforces a safety margin, $\epsilon_{\text{safe}}$, from world obstacles, $O_j$, by applying a hinge loss on the signed distance, $d_s$, of the robot's swept volume:
\begin{equation} \mathcal{C}_{\text{collision}} = \sum_{t,j} w_c \cdot \max(0, \epsilon_{\text{safe}} - d_s(q_t, q_{t+1}, O_j))^2. \end{equation}

We implement the optimization using PyRoki~\cite{kim_pyroki_2025} and Jax.
The cost terms are weighted as follows.
The joint limit penalty is set to a high value of $w_l=100.0$ to act as a hard constraint.
The smoothness weight is $w_s=10.0$, the collision penalty weight is $w_c=15.0$, and a small regularization is applied with a rest pose weight of $w_r=0.1$.

\newpage
\section{Experiments}
We show more visualizations of the real-world manipulation experiments in \cref{fig:real_world_experiments_appendix}.
\shortname is versatile and supports cross-embodiment manipulation, which we use to manipulate rigid, deformable, and articulated objects using tabletop and mobile manipulator.
It is also viewpoint-agnostic and can be deployed on a novel platform after performing hand-eye calibration.

\end{appendices}

\end{document}